\documentclass[twoside,11pt]{article}
\usepackage{jmlr2e}
\newdimen\pHeight
\newdimen\pLower
\newdimen\pLineWidth
\newdimen\pKern
\newdimen\pIR
\newsavebox{\Cbox}
\newsavebox{\vertCmplx}
\newdimen\Cheight
\newdimen\Cwidth
\sbox{\Cbox}{\rm C}
\Cheight=\ht\Cbox
\Cwidth=\wd\Cbox
\advance\Cheight by \pHeight
\sbox{\vertCmplx}{\rule[\pLower]{\pLineWidth}{\Cheight}}
\sbox{\Cbox}{\usebox{\Cbox}\kern\pKern\usebox{\vertCmplx}}
\wd\Cbox=\Cwidth
\def\Re{\mathbb{R}}

\def\Nat{{\rm I\kern\pIR N}}

\def\Var{{\bf Var}}

\def\A{{\mathcal{A}}}

\def\D{{\mathcal{D}}}
\def\E{{\mathcal{E}}}
\def\F{{\mathcal{F}}}
\def\G{{\mathcal{G}}}

\def\I{{\mathcal{I}}}

\def\L{{\mathcal{L}}}
\def\M{{\mathcal{M}}}

\def\P{{\mathcal{P}}}

\def\R{{\mathcal{R}}}
\def\S{{\mathcal{S}}}

\def\Z{{\mathcal{Z}}}

\newcommand{\ra}{\rightarrow}
\newcommand{\beq}{\begin{equation}}
\newcommand{\eeq}{\end{equation}}
\newcommand{\beqa}{\begin{eqnarray}}
\newcommand{\eeqa}{\end{eqnarray}}
\newcommand{\beqan}{\begin{eqnarray*}}
\newcommand{\eeqan}{\end{eqnarray*}}
\newcommand{\ben}{\begin{eqnarray*}}
\newcommand{\een}{\end{eqnarray*}}

\usepackage{graphicx}
\usepackage{amsmath,amsfonts,amssymb,bm,eucal}
\usepackage{algorithm,algorithmic}
\usepackage[usenames,dvipsnames]{xcolor}
\usepackage{xspace}
\usepackage{varwidth}
\usepackage{enumitem}


\newcounter{parentnumber}

\def\Tl{T^{(\l)}}

\jmlrheading{}{}{}{}{}{Richard S. Sutton, A. Rupam Mahmood \& Martha White}

\ShortHeadings{An Emphatic Approach to Off-policy TD Learning}{Sutton, Mahmood \& White}
\firstpageno{1}

\title{An Emphatic Approach to the Problem \\ of Off-policy Temporal-Difference Learning}

\author{\name Richard S. Sutton{\email sutton@cs.ualberta.ca}\\
              A. Rupam Mahmood{\email ashique@cs.ualberta.ca}\\
              Martha White\email whitem@cs.ualberta.ca\\
       \addr Reinforcement Learning and Artificial Intelligence Laboratory\\
       Department of Computing Science, University of Alberta\\
       Edmonton, Alberta, Canada T6G\,2E8}

\editor{}

\allowdisplaybreaks[1]
\def\beq{\begin{equation}}
\def\eeq{\end{equation}}

\def\a{\alpha}

\def\e{\epsilon}
\def\g{\gamma}
\def\i{I}
\def\l{\lambda}
\def\tr{^\top}
\def\th{{\bm\theta}}
\def\ph{{\bm\phi}}
\def\thet{\theta}
\def\phix{\phi}
\def\E#1{{\mathbb E}\!\left[#1\right]}

\def\Var#1{\mathrm{Var}\!\left[#1\right]}

\def\Epi#1{{\mathbb E}_\pi\!\left[#1\right]}
\def\Emu#1{{\mathbb E}_\mu\!\left[#1\right]}
\def\Pr#1{{\mathbb P}\!\left\{#1\right\}}

\def\CP#1#2{{\mathbb P}\!\left\{#1\middle\vert{#2}\right\}}
\def\CE#1#2{{\mathbb E}\!\left[#1\middle\vert{#2}\right]}
\def\CEmu#1#2{{\mathbb E}_\mu\!\left[#1\middle\vert{#2}\right]}
\def\CEmubg#1#2{{\mathbb E}_\mu\bigl[#1\!\bigm\vert\!{#2}\bigr]}
\def\CEpi#1#2{{\mathbb E}_\pi\!\left[#1\middle\vert{#2}\right]}

\def\la{($\lambda$)\xspace}
\def\ra{\rightarrow}
\def\vpi{v_{\pi}}
\def\MSVE{\mbox{\sc MSVE}\xspace}
\def\Gl{G^{\l}}
\def\Glr{G^{\l\rho}}
\def\Tl{T^{\l}}

\def\thtwoth{$\thet\!\ra\!2\thet$\xspace}
\def\so{\bar s} 
\def\ao{\bar a} 

\def\lef{{\sf left}\xspace}
\def\righ{{\sf right}\xspace}
\newcommand{\shortminus}{\scalebox{0.75}[1.0]{\( - \)}}
\def\sm{\shortminus}

\def\bb{{\bf b}}

\def\dd{{\bf d}}
\def\ee{{\bf e}}
\def\ff{{\bf f}}

\def\ii{{\bf i}}
\def\mm{{\bf m}}
\def\rr{{\bf r}}
\def\rpi{\rr_\pi}

\def\rl{\rr_\pi^{\l}}
\def\vv{{\bf v}}

\def\yy{{\bf y}}

\def\1{{\bf 1}}

\def\A{{\bf A}}

\def\D{{\bf D_{\!\mu}}}
\def\Dpi{{\bf D_{\pi}}}
\def\Em{{\bf E}}
\def\F{{\bf F}}
\def\G{{\bf \Gamma}}
\def\L{{\bf \Lambda}}
\def\I{{\bf I}}
\def\M{{\bf M}}
\def\P{{\bf P}}
\def\Pg{\P_{\!\g}}
\def\Pl{{\P_{\pi}^{\l}}}
\def\Ppi{\P_{\!\pi}}

\def\R{{\bf R}}
\def\PH{{\bm\Phi}}
\def\S{{\bf S}}

\def\Z{Z}

\def\SS{{\cal S}}
\def\AA{{\cal A}}

\begin{document}

\maketitle

\begin{abstract}%
In this paper we introduce the idea of improving the performance of parametric temporal-difference (TD) learning algorithms by selectively emphasizing or de-emphasizing their updates on different time steps. In particular, we show that varying the emphasis of linear TD\la's updates in a particular way causes its expected update to become stable under off-policy training. The only prior model-free TD methods to achieve this with per-step computation linear in the number of function approximation parameters are the gradient-TD family of methods including TDC, GTD\la, and GQ\la. Compared to these methods, our \emph{emphatic TD\la} is simpler and easier to use; it has only one learned parameter vector and one step-size parameter. 
Our treatment includes general state-dependent discounting and bootstrapping functions, and a way of specifying varying degrees of interest in accurately valuing different states. 
\end{abstract}

\begin{keywords}
temporal-difference learning, off-policy training, function approximation, convergence, stability
\end{keywords}

\section{Parametric Temporal-Difference Learning}

Temporal-difference (TD) learning is perhaps the most important idea to come out of the field of reinforcement learning. The problem it solves is that of efficiently learning to make a sequence of long-term predictions about how a dynamical system will evolve over time. The key idea is to use the change (temporal difference) from one prediction to the next as an error in the earlier prediction. For example, if you are predicting on each day what the stock-market index will be at the end of the year, and events lead you one day to make a much lower prediction, then a TD method would infer that the predictions made prior to the drop were probably too high; it would adjust the parameters of its prediction function so as to make lower predictions for similar situations in the future. This approach contrasts with conventional approaches to prediction, which wait until the end of the year when the final stock-market index is known before adjusting any parameters, or else make only short-term (e.g., one-day) predictions and then iterate them to produce a year-end prediction. The TD approach is more convenient computationally because it requires less memory and because its computations are spread out uniformly over the year (rather than being bunched together all at the end of the year). A less obvious advantage of the TD approach is that it often produces statistically more accurate answers than conventional approaches (Sutton 1988).

Parametric temporal-difference learning was first studied as the key ``learning by generalization" algorithm in Samuel's (1959) checker player. Sutton (1988) introduced the TD\la algorithm and proved convergence in the mean of episodic linear TD(0), the simplest parametric TD method. The potential power of parametric TD learning was convincingly demonstrated by Tesauro (1992, 1995) when he applied TD\la combined with neural networks and self play to obtain ultimately the world's best backgammon player. Dayan (1992) proved convergence in expected value of episodic linear TD\la for all $\l\in[0,1]$, and Tsitsiklis and Van Roy (1997) proved convergence with probability one of discounted continuing linear TD\la. Watkins (1989) extended TD learning to control in the form of Q-learning and proved its convergence in the tabular case (without function approximation, Watkins \& Dayan 1992), while Rummery (1995) extended TD learning to control in an on-policy form as the Sarsa\la algorithm. Bradtke and Barto (1996), Boyan (1999), and Nedic and Bertsekas (2003) extended linear TD learning to a least-squares form called LSTD\la. Parametric TD methods have also been developed as models of animal learning (e.g., Sutton \& Barto 1990, Klopf 1988, Ludvig, Sutton \& Kehoe 2012) and as models of the brain's reward systems (Schultz, Dayan \& Montague 1997), where they have been particularly influential (e.g., Niv \& Schoenbaum 2008, O'Doherty 2012). Sutton (2009, 2012) has suggested that parametric TD methods could be key not just to learning about reward, but to the learning of world knowledge generally, and to perceptual learning.
Extensive analysis of parametric TD learning as stochastic approximation is provided by Bertsekas (2012, Chapter 6) and Bertsekas and Tsitsiklis (1996).

Within reinforcement learning, TD learning is typically used to learn approximations to the value function of a Markov decision process (MDP). Here the value of a state $s$, denoted $\vpi(s)$, is defined as the sum of the expected long-term discounted rewards that will be received if the process starts in $s$ and subsequently takes actions as specified by the decision-making policy $\pi$, called the \emph{target policy}. If there are a small number of states, then it may be practical to approximate the function $\vpi$ by a table, but more generally a parametric form is used, such as a polynomial, multi-layer neural network, or linear mapping. Also key is the source of the data, in particular, the policy used to interact with the MDP. If the data is obtained while following the target policy $\pi$, then good convergence results are available for linear function approximation. This case is called \emph{on-policy} learning because learning occurs while ``on" the policy being learned about. In the alternative, \emph{off-policy} case, one seeks to learn about $\vpi$ while behaving (selecting actions) according to a different policy called the \emph{behavior policy}, which we denote by $\mu$. Baird (1995) showed definitively that parametric TD learning was much less robust in the off-policy case by exhibiting counterexamples for which both linear TD(0) and linear Q-learning had unstable expected updates and, as a result, the parameters of their linear function approximation diverged to infinity. This is a serious limitation, as the off-policy aspect is key to Q-learning (perhaps the single most popular reinforcement learning algorithm), to learning from historical data and from demonstrations, and to the idea of using TD learning for perception and world knowledge.

Over the years, several different approaches have been taken to solving the problem of off-policy learning. Baird (1995) proposed an approach based on gradient descent in the Bellman error for general parametric function approximation that has the desired computational properties, but which requires access to the MDP for double sampling and which in practice often learns slowly. Gordon (1995, 1996) proposed restricting attention to function approximators that are averagers, but this does not seem to be possible without storing many of the training examples, which would defeat the primary strength that we seek to obtain from parametric function approximation. The LSTD\la method was always relatively robust to off-policy training (e.g., Lagoudakis \& Parr 2003, Yu 2010, Mahmood, van Hasselt \& Sutton 2014), but its per-step computational complexity is quadratic in the number of parameters of the function approximator, as opposed to the linear complexity of TD\la and the other methods. Perhaps the most successful approach to date is the gradient-TD approach (e.g., Maei 2011, Sutton et al.\ 2009, Maei et al.\ 2010), including hybrid methods such as HTD (Hackman 2012). Gradient-TD methods are of linear complexity and guaranteed to converge for appropriately chosen step-size parameters but are more complex than TD\la because they require a second auxiliary set of parameters with a second step size that must be set in a problem-dependent way for good performance. The studies by White (in preparation), Geist and Scherrer (2014), and Dann, Neumann, and Peters (2014) are the most extensive empirical explorations of gradient-TD and related methods to date.

In this paper we explore a new approach to solving the problem of off-policy TD learning with function approximation. The approach has novel elements but is similar to that developed by Precup, Sutton, and Dasgupta in 2001. They proposed to use importance sampling to reweight the updates of linear TD\la, emphasizing or de-emphasizing states as they were encountered, and thereby create a weighting equivalent to the stationary distribution under the target policy, from which the results of Tsitsiklis and Van Roy (1997) would apply and guarantee convergence. As we discuss later, this approach has very high variance and was eventually abandoned in favor of the gradient-TD approach. The new approach we explore in this paper is similar in that it also varies emphasis so as to reweight the distribution of linear TD\la updates, but to a different goal. The new goal is to create a weighting equivalent to the \emph{followon distribution} for the target policy started in the stationary distribution of the behavior policy. The followon distribution weights states according to how often they would occur prior to termination by discounting \emph{if the target policy was followed}. 

Our main result is to prove that varying emphasis according to the followon distribution produces a new version of linear TD\la, \emph{called emphatic TD\la}, that is stable under general off-policy training. By ``stable" we mean that the expected update over the ergodic distribution (Tsitsiklis \& Van Roy 1997) is a contraction, involving a positive definite matrix. We concentrate on stability in this paper because it is a prerequisite for full convergence of the stochastic algorithm.
Demonstrations that the linear TD\la is not stable under off-policy training have been the focus of previous counterexamples (Baird 1995, Tsitsiklis \& Van Roy 1996, 1997, see Sutton \& Barto 1998). Substantial additional theoretical machinery would be required for a full convergence proof. Recent work by Yu (in preparation) builds on our stability result to prove that the emphatic TD\la converges with probability one.

In this paper we first treat the simplest algorithm for which the difficulties of off-policy temporal-difference (TD) learning arise---the TD(0) algorithm with linear function approximation. We examine the conditions under which the expected update of on-policy TD(0) is stable, then why those conditions do not apply under off-policy training, and finally how they can be recovered for off-policy training using established importance-sampling methods together with the emphasis idea.
After introducing the basic idea of emphatic algorithms using the special case of TD(0), we then develop the general case. In particular, we consider a case with general state-dependent discounting and bootstrapping functions, and with a user-specified allocation of function approximation resources. 
Our new theoretical results and the emphatic TD\la algorithm are presented fully for this general case. Empirical examples elucidating the main theoretical results are presented in the last section prior to the conclusion.

\section{On-policy Stability of TD(0)}

To begin, let us review the conditions for stability of conventional TD\la under on-policy training with data from a continuing finite Markov decision process. Consider the simplest function approximation case, that of linear TD\la with $\l=0$ and constant discount-rate parameter $\g\in[0,1)$. 
Conventional linear TD(0) is defined by the following update to the parameter vector $\th_t\in\Re^n$, made at each of a sequence of time steps $t=0,1,2,\ldots$, on transition from state $S_t\in\SS$ to state $S_{t+1}\in\SS$, taking action $A_t\in\AA$ and receiving reward $R_{t+1}\in\Re$:
\begin{align} \label{eq:TD0}
  \th_{t+1} \doteq \th_t + \a\left(R_{t+1}+\g\th_t\tr\ph(S_{t+1})-\th_t\tr\ph(S_t)\right)\ph(S_t), 
\end{align}
where $\a>0$ is a step-size parameter, and $\ph(s)\in\Re^n$ is the feature vector corresponding to state $s$. The notation ``$\doteq$" indicates an equality by definition rather than one that follows from previous definitions.
In on-policy training, the actions are chosen according to a target policy $\pi:\AA\times\SS\ra[0,1]$, where $\pi(a|s) \doteq \CP{A_t\!=\!a}{S_t\!=\!s}$. 
The state and action sets $\SS$ and $\AA$ are assumed to be finite, but the number of states is assumed much larger than the number of learned parameters, $|\SS| \doteq N \gg n$, so that function approximation is necessary. We use linear function approximation, in which
the inner product of the parameter vector and the feature vector for a state is meant to be an approximation to the value of that state:
\beq
\th_t\tr\ph(s) \approx \vpi(s) \doteq \CEpi{G_t}{S_t\!=\!s}, \label{eq:vpi}
\eeq
where $\Epi{\cdot}$ denotes an expectation conditional on all actions being selected according to $\pi$, and $G_t$, the \emph{return} at time $t$, is defined by
\begin{align}
G_t &\doteq R_{t+1} + \g R_{t+2} + \g^2 R_{t+3} + \cdots. \label{eq:G}
\end{align}

The TD(0) update \eqref{eq:TD0} can be rewritten to make the stability issues more transparent:
\begin{align}
  \th_{t+1} 
  &= \th_t + \a\Big(\underbrace{R_{t+1}\ph(S_t)}_{\bb_t\in\Re^n} - \underbrace{\ph(S_t)\left(\ph(S_t)-\g\ph(S_{t+1})\right)\tr}_{\A_t\in\Re^{n\times n}}\th_t\Big) \nonumber\\
  &= \th_t + \a(\bb_t - \A_t\th_t)\label{eq:Abt}\\  
  &= (\I-\a\A_t)\th_t + \a\bb_t. \nonumber
\end{align}
The matrix $\A_t$ multiplies the parameter $\th_t$ and is thereby critical to the stability of the iteration. To develop intuition, consider the special case in which $\A_t$ is a diagonal matrix. If any of the diagonal elements are negative, then the corresponding diagonal element of $\I-\a\A_t$ will be greater than one, and the corresponding
component of $\th_t$ will be amplified, which will lead to divergence if continued. (The second term ($\a\bb_t$) does not affect the stability of the iteration.) On the other hand, if the diagonal elements of $\A_t$ are all positive, then $\a$ can be chosen smaller than one over the largest of them, such that $\I-\a\A_t$ is diagonal with all diagonal elements between 0 and 1. In this case the first term of the update tends to shrink $\th_t$, and stability is assured. In general, $\th_t$ will be reduced toward zero whenever $\A_t$ is positive definite.\footnote{A real matrix $\A$ is defined to be \emph{positive definite} in this paper iff $\yy\tr\!\A\yy>0$ for any real vector $\yy\not=\bm 0$.}

In actuality, however, $\A_t$ and $\bb_t$ are random variables that vary from step to step, in which case stability is determined by the steady-state expectation, $\lim_{t\ra\infty}\E{\A_t}$. In our setting, after an  initial transient, states will be visited according to the steady-state distribution under $\pi$. We represent this distribution by a vector $\dd_\pi$, each component of which gives the limiting probability of being in a particular state\footnote{Here and throughout the paper we use brackets with subscripts to denote the individual elements of vectors and matrices.} $[\dd_\pi]_s  \doteq d_\pi(s)\doteq \lim_{t\ra\infty} \Pr{S_t\!=\!s}$, which we assume exists and is positive at all states (any states not visited with nonzero probability can be removed from the problem). The special property of the steady-state distribution is that once the process is in it, it remains in it. Let $\P_\pi$ denote the $N\times N$ matrix of transition probabilities $[\P_\pi]_{ij} \doteq  \sum_a \pi(a|i)p(j|i,a)$ where $p(j|i,a)\doteq \CP{S_{t+1}\!=\!j}{S_t\!=\!i,A_t\!=\!a}$. Then the special property of $\dd_\pi$ is that 
\beq
\P_\pi\tr\dd_\pi = \dd_\pi. \label{eq:stationary}
\eeq

\def\RP{\mbox{\rm Re}}
\def\thstar{\bar\th}

Consider any stochastic algorithm of the form \eqref{eq:Abt}, and let $\A\doteq\lim_{t\ra\infty}\E{\A_t}$ and $\bb\doteq\lim_{t\ra\infty}\E{\bb_t}$. 
We define the stochastic algorithm to be \emph{stable} if and only if the corresponding deterministic algorithm, 
\begin{align}
  \bar\th_{t+1} \doteq \bar\th_t + \a(\bb - \A\bar\th_t),  \label{eq:Eth}
\end{align}
is convergent to a unique fixed point independent of the initial $\bar\th_0$. This will occur iff the $\A$ matrix has a full set of eigenvalues all of whose real parts are positive.
If a stochastic algorithm is stable and $\a$ is reduced according to an appropriate schedule, then its parameter vector may converge with probability one. 
However, in this paper we focus only on stability as a prerequisite for convergence, leaving convergence itself to future work.
If the stochastic algorithm converges, it is to a fixed point $\thstar$ of the deterministic algorithm, at which $\A\thstar=\bb$, or $\thstar=\A^{-1}\bb$. (Stability assures existence of the inverse.)
In this paper we focus on establishing stability by proving that $\A$ is positive definite. From definiteness it immediately follows that $\A$ has a full set of eigenvectors (because $\yy\tr\A\yy>0, \forall \yy\not=\bm 0$) and that the corresponding eigenvalues all have real parts.\footnote{To see the latter, let $\RP(x)$ denote the real part of a complex number $x$, and let $\yy^*$ denotes the conjugate transpose of a complex vector $\yy$. Then, for any eigenvalue--eigenvector pair $\l, \yy$: $0< \RP(\yy^*\A\yy)=\RP(\yy^*\l\yy)=\RP(\l) \yy^*\yy \implies 0<\RP(\l)$.}

Now let us return to analyzing on-policy TD(0). Its $\A$ matrix is
\begin{align*}
\A = \lim_{t\ra\infty}\E{\A_t} 
 &= \lim_{t\ra\infty}\Epi{\ph(S_t)\left(\ph(S_t)-\g\ph(S_{t+1})\right)\tr} \\
 &= \sum_s d_\pi(s)\,\ph(s)\!\left(\ph(s)-\g\sum_{s'}[\P_\pi]_{ss'}\ph(s')\right)\tr \\ 
 &= \PH\tr\Dpi(\I-\g\P_\pi)\PH, 
\end{align*}
where $\PH$ is the $N\times n$ matrix with the $\ph(s)$ as its rows, and $\Dpi$ is the $N\times N$ diagonal matrix with $\dd_\pi$ on its diagonal.
This $\A$ matrix is typical of those we consider in this paper in that it consists of $\PH\tr$ and $\PH$ wrapped around a distinctive $N\times N$ matrix that varies with the algorithm and the setting, and which we call the \emph{key matrix}. An $\A$ matrix of this form will be positive definite whenever the corresponding key matrix is positive definite.\footnote{Strictly speaking, positive definiteness of the key matrix assures only that $\A$ is positive \emph{semi-}definite, because it is possible that $\PH\yy=\bm 0$ for some $\yy\not=\bm 0$, in which case $\yy\tr \A\yy$ will be zero as well. To rule this out, we assume, as is commonly done, that the columns of $\PH$ are linearly independent (i.e., that the features are not redundant), and thus that $\PH\yy=\bm 0$ only if $\yy=\bm 0$. If this were not true, then convergence (if it occurs) may not be to a unique $\th_\infty$, but rather to a subspace of parameter vectors all of which produce the same approximate value function.}
In this case the key matrix is $\Dpi(\I-\g \Ppi)$. 

For a key matrix of this type, positive definiteness is assured if all of its columns sum to a nonnegative number. This was shown by Sutton (1988, p.~27) based on two previously established theorems. One theorem says that any matrix $\M$ is positive definite if and only if the symmetric matrix $\S = \M + \M\tr$  is positive definite (Sutton 1988, appendix). The second theorem says that any symmetric real matrix $\S$ is positive definite if all of its diagonal entries are positive and greater than the sum of the corresponding off-diagonal entries (Varga 1962, p.~23). For our key matrix, $\Dpi(\I-\g \Ppi)$, the diagonal entries are positive and the off-diagonal entries are negative, so all we have to show is that each row sum plus the corresponding column sum is positive. The row sums are all positive because $\Ppi$ is a stochastic matrix and $\g<1$. Thus it only remains to show that the column sums are nonnegative. Note that the row vector of the column sums of any matrix $\M$ can be written as $\1\tr\M$, where $\1$ is the column vector with all components equal to 1. The column sums of our key matrix, then, are:
\begin{align*}
 \1\tr\Dpi(\I-\g\Ppi)
  &= \dd_\pi\tr(\I-\g \Ppi) \\
  &= \dd_\pi\tr-\g \dd_\pi\tr\Ppi) \\
  &= \dd_\pi\tr-\g \dd_\pi\tr \tag{by \eqref{eq:stationary}}\\
  &= (1-\g)\dd_\pi, 
\end{align*}
all components of which are positive.
Thus, the key matrix and its $\A$ matrix are positive definite, and on-policy TD(0) is stable. Additional conditions and a schedule for reducing $\a$ over time (as in Tsitsiklis and Van Roy 1997) are needed to prove convergence with probability one, $\th_\infty=\A^{-1}\bb$, but the analysis above includes the most important steps that vary from algorithm to algorithm.

\section{Instability of Off-policy TD(0)}

Before developing the off-policy setting in detail, it is useful to understand informally why TD(0) is susceptible to instability. TD learning involves learning an estimate from an estimate, which can be problematic if there is generalization between the two estimates. For example, suppose there is a transition between two states with the same feature representation except that the second is twice as big:

\vspace{10pt}
\centerline{\includegraphics[height=0.3in]{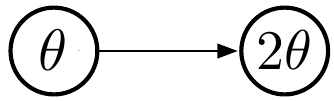}}
\vspace*{1pt}

\noindent
where here $\thet$ and $2\thet$ are the estimated values of the two states---that is, their feature representations are a single feature that is 1 for the first state and 2 for the second (Tsitsiklis \& Van Roy 1996, 1997). Now suppose that $\thet$ is 10 and the reward on the transition is 0. The transition is then from a state valued at 10 to a state valued at 20. If $\g$ is near 1 and $\a$ is 0.1, then $\thet$ will be increased to approximately $11$. But then the next time the transition occurs there will be an even bigger increase in value, from 11 to 22, and a bigger increase in $\thet$, to approximately $12.1$. If this transition is experienced repeatedly on its own, then the system is unstable and the parameter increases without bound---it diverges. We call this the \thtwoth problem.

In on-policy learning, repeatedly experiencing just this single problematic transition  cannot happen, because, after the highly-valued $2\thet$ state has been entered, it must then be exited. The transition from it will either be to a lesser or equally-valued state, in which case $\thet$ will be significantly decreased, or to an even higher-valued state which in turn must be followed by an even larger decrease in its estimated value or a still higher-valued state. Eventually, the promise of high value must be made good in the form of a high reward, or else estimates will be decreased, and this ultimately constrains $\thet$ and forces stability and convergence. In the off-policy case, however, if there is a deviation from the target policy then the promise is excused and need never be fulfilled. Later in this section we present a complete example of how the \thtwoth problem can cause instability and divergence under off-policy training.

%

With these intuitions, we now detail our off-policy setting. As in the on-policy case, the data is a single, infinite-length trajectory of actions, rewards, and feature vectors generated by a continuing finite Markov decision process. The difference is that the actions are selected not according to the target policy $\pi$, but according to a different \emph{behavior policy} $\mu:\AA\times\SS\ra[0,1]$, yet still we seek to estimate state values under $\pi$ (as in \eqref{eq:vpi}). 
Of course, it would be impossible to estimate the values under $\pi$ if the actions that $\pi$ would take were never taken by $\mu$ and their consequences were never observed. Thus we assume that $\mu(a|s)>0$ for every state and action for which $\pi(a|s)>0$. This is called the assumption of \emph{coverage}. It is trivially satisfied by any $\e$-greedy or soft behavior policy.
As before we assume that there is a stationary distribution $d_\mu(s)\doteq\lim_{t\ra\infty}\Pr{S_t\!=\!s}>0, \forall s\in\SS$, with corresponding $N$-vector $\dd_\mu$.

Even if there is coverage, the behavior policy will choose actions with proportions different from the target policy. For example, some actions taken by $\mu$ might never be chosen by $\pi$. To address this, we use importance sampling to correct for the relative probability of taking the action actually taken, $A_t$, in the state actually encountered, $S_t$, under the target and behavior policies:
\[
 \rho_t \doteq  \frac{\pi(A_t|S_t)}{\mu(A_t|S_t)}. 
\]
This quantity is called the \emph{importance sampling ratio}
at time $t$. Note that its expected value is one:
\[
 \CEmu{\rho_t}{S_t\!=\!s} = \sum_a \mu(a|s) \frac{\pi(a|s)}{\mu(a|s)} = \sum_a \pi(a|s) = 1.
\]
The ratio will be exactly one only on time steps on which the action probabilities for the two policies are exactly the same; these time steps can be treated the same as in the on-policy case. On other time steps the ratio will be greater or less than one depending on whether the action taken was more or less likely under the target policy than under the behavior policy, and some kind of correction is needed. 

In general, for any random variable $\Z_{t+1}$ dependent on $S_t$, $A_t$ and $S_{t+1}$, we can recover its expectation under the target policy by multiplying by the importance sampling ratio:
\begin{align}
 \CEmu{\rho_t \Z_{t+1}}{S_t\!=\!s} 
 &= \sum_a \mu(a|s) \frac{\pi(a|s)}{\mu(a|s)} \Z_{t+1} \nonumber\\
 &= \sum_a \pi(a|s) \Z_{t+1} \nonumber\\
 &= \CEpi{\Z_{t+1}}{S_t\!=\!s}, ~~~~\forall s\in\SS. \label{eq:Eshift}
\end{align}
We can use this fact to begin to adapt TD(0) for off-policy learning (Precup, Sutton \& Singh 2000). We simply multiply the whole TD(0) update \eqref{eq:TD0}
by $\rho_t$:
\begin{align} \label{eq:offTD0}
  \th_{t+1} &\doteq  \th_t + \rho_t\,\a\left(R_{t+1}+\g\th_t\tr\ph_{t+1}-\th_t\tr\ph_t\right)\ph_t\\
  &= \th_t + \a\Big(\underbrace{\rho_t R_{t+1}\ph_t}_{\bb_t} - \underbrace{\rho_t\ph_t\left(\ph_t-\g\ph_{t+1}\right)\tr}_{\A_t}\th_t\Big), \nonumber
  \end{align}
where here we have used the shorthand $\ph_t\doteq \ph(S_t)$. Note that if the action taken at time $t$ is never taken under the target policy in that state, then $\rho_t=0$ and there is no update on that step, as desired.
We call this algorithm \emph{off-policy TD(0)}. 

Off-policy TD(0)'s $\A$ matrix is
\begin{align}
\A = \lim_{t\ra\infty}\E{\A_t} 
 &= \lim_{t\ra\infty}\Emu{\rho_t\ph_t\left(\ph_t-\g\ph_{t+1}\right)\tr} \nonumber\\
 &= \sum_s d_\mu(s)\CEmu{\rho_k\ph_k\left(\ph_k-\g\ph_{k+1}\right)\tr}{S_k=s} \nonumber\\ 
 &= \sum_s d_\mu(s)\CEpi{\ph_k\left(\ph_k-\g\ph_{k+1}\right)\tr}{S_k=s} \tag{by \eqref{eq:Eshift}}\\ 
 &= \sum_s d_\mu(s)\,\ph(s)\!\left(\ph(s)-\g\sum_{s'}[\P_\pi]_{ss'}\ph(s')\right)\tr \nonumber\\ 
 &= \PH\tr\D(\I-\g\P_\pi)\PH,  \nonumber
\end{align}
where $\D$ is the $N\times N$ diagonal matrix with the stationary distribution $\dd_\mu$ on its diagonal. Thus, the key matrix that must be positive definite is $\D(\I-\g\Ppi)$ and, unlike in the on-policy case, the distribution and the transition probabilities do not match. We do not have an analog of \eqref{eq:stationary}, $\Ppi\tr\dd_\mu\not=\dd_\mu$, and in fact the column sums may be negative and the matrix not positive definite, in which case divergence of the parameter is likely.

\begin{figure}[b]
\vspace*{6pt}
\centerline{\includegraphics[height=0.4in]{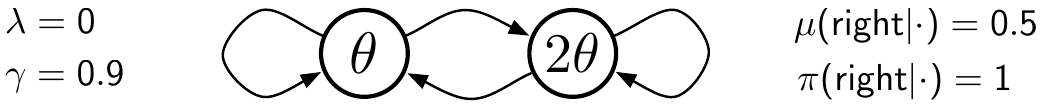}}
\caption{\thtwoth example without a terminal state.}
\label{fig:th2thcontinuing}
\end{figure}

A simple \thtwoth example of divergence that fits the setting in this section is shown in Figure~\ref{fig:th2thcontinuing}.
From each state there are two actions, \lef and \righ, which take the process to the left or right states. All the rewards are zero. As before, there is a single parameter $\thet$ and the single feature is 1 and 2 in the two states such that the approximate values are $\thet$ and $2\thet$ as shown. The behavior policy is to go \lef and \righ with equal probability from both states, such that equal time is spent on average in both states, $\dd_\mu=(0.5, 0.5)\tr$. The target policy is to go \righ in both states. We seek to learn the value from each state given that the \righ action is continually taken. 
The transition probability matrix for this example is:
\[ \Ppi = \begin{bmatrix} 0 & 1 \\ 0 & 1 \end{bmatrix}. \]
The key matrix is
\beq
\D(\I-\g\Ppi) 
= \begin{bmatrix} 0.5 & 0 \\ 0 & 0.5 \end{bmatrix}
\times \begin{bmatrix} 1 & \sm0.9 \\ 0 & 0.1 \end{bmatrix}
= \begin{bmatrix} 0.5 & \sm0.45 \\ 0 & 0.05 \end{bmatrix}. \label{eq:keyoff}
\eeq
We can see an immediate indication that the key matrix may not be positive definite in that the second column sums to a negative number. 
More definitively, one can show that it is not positive definite by multiplying it on both sides by $\yy=\PH=(1, 2)\tr$:
\[
\PH\tr\D(\I-\g\Ppi)\PH = \begin{bmatrix} 1 & 2 \end{bmatrix}
\times\begin{bmatrix} 0.5 & \sm0.45 \\ 0 & 0.05 \end{bmatrix}
\times\begin{bmatrix} 1 \\ 2 \end{bmatrix}
= \begin{bmatrix} 1 & 2 \end{bmatrix}
\times\begin{bmatrix} \sm0.4 \\ 0.1 \end{bmatrix}
= -0.2.
\]
That this is negative means that the key matrix is not positive definite. We have also calculated here the $\A$ matrix; it is this negative scalar, $\A=-0.2$. Clearly, this expected update and algorithm are not stable.

It is also easy to see the instability of this example more directly, without matrices. We know that only transitions under the \righ action cause updates, as $\rho_t$ will be zero for the others. Assume for concreteness that initially $\thet_t=10$ and that $\a=0.1$. On a \righ transition from the first state the update will be
\begin{align*}
\thet_{t+1} &= \thet_t+\rho_t\a\left(R_{t+1}+\g\th_t\tr\ph_{t+1}-\th_t\tr\ph_t\right)\ph_t\\
             &= 10 + 2\cdot 0.1 \left(0+0.9\cdot 10\cdot 2 - 10\cdot 1\right) 1\\
             &= 10 + 1.6,  
\end{align*}
whereas, on a \righ transition from the second state the update will be
\begin{align*}
\thet_{t+1} &= \thet_t+\rho_t\a\left(R_{t+1}+\g\th_t\tr\ph_{t+1}-\th_t\tr\ph_t\right)\ph_t\\
             &= 10 + 2\cdot 0.1 \left(0+0.9\cdot 10\cdot 2 - 10\cdot 2\right) 2\\
             &= 10 -0.8.
\end{align*}
These two transitions occur equally often, so the net change will be positive.
That is, $\thet$ will increase, moving farther from its correct value, zero.  Everything is linear in $\thet$, so the next time around, with a larger starting $\thet$, the increase in $\thet$ will be larger still, and divergence occurs. A smaller value of $\a$ would not prevent divergence, only reduce its rate.

\section{Off-policy Stability of Emphatic TD(0)}

The deep reason for the difficulty of off-policy learning is that the behavior policy may take the process to a distribution of states different from that which would be encountered under the target policy, yet the states might appear to be the same or similar because of function approximation. Earlier work by Precup, Sutton and Dasgupta (2001) attempted to completely correct for the different state distribution using importance sampling ratios to reweight the states encountered. It is theoretically possible to convert the state weighting from $d_\mu$ to $d_\pi$ using the product of all importance sampling ratios from time 0, but in practice this approach has extremely high variance and is infeasible for the continuing (non-episodic) case. It works in theory because after converting the weighting the key matrix is $\Dpi(\I-\g\Ppi)$ again, which we know to be positive definite. 

Most subsequent works abandoned the idea of completely correcting for the state distribution. For example, the work on gradient-TD methods (e.g., Sutton et al.\ 2009, Maei 2011) seeks to minimize the mean-squared projected Bellman error weighted by $d_\mu$.
We call this an \emph{excursion} setting because we can think of the contemplated switch to the target policy as an excursion from the steady-state distribution of the behavior policy, $d_\mu$. The excursions would start from $d_\mu$ and then follow $\pi$ until termination, followed by a resumption of $\mu$ and thus a gradual return to $d_\mu$.
Of course these excursions never actually occur during off-policy learning, they are just contemplated, and thus the state distribution in fact never leaves $d_\mu$. It is the excursion view that we take in this paper, but still we use techniques similar to those introduced by Precup et al.\ (2001) to determine an emphasis weighting that corrects for the state distribution, only toward a different goal.\footnote{Kolter (2011) also suggested adapting the distribution of states at which updates are made to improve convergence and solution quality, but did not provide a linear-complexity algorithm.}

The excursion notion suggests a different weighting of TD(0) updates. We consider that at every time step we are beginning a new contemplated excursion from the current state. 
The excursion thus would begin in a state sampled from $d_\mu$. If an excursion started it would pass through a sequence of subsequent states and actions prior to termination. Some of the actions that are actually taken (under $\mu$) are relatively likely to occur under the target policy as compared to the behavior policy, while others are relatively unlikely; the corresponding states can be appropriately reweighted based on importance sampling ratios. Thus, there will still be a product of importance sampling ratios, but only since the beginning of the excursion, and the variance will also be tamped down by the discounting; the variance will be much less than in the earlier approach.
In the simplest case of an off-policy emphatic algorithm, the update at time $t$ is emphasized or de-emphasized proportional to a new scalar variable $F_t$, defined by $F_0=1$ and
\beq
 F_{t} \doteq  \g\rho_{t-1}F_{t-1} + 1,  ~~~~~\forall t>0, 
\eeq
which we call the \emph{followon trace}.
Specifically, we define \emph{emphatic TD(0)} by the following update:
\begin{align} \label{eq:ETD0}
  \th_{t+1} &\doteq  \th_t + \a F_t\rho_t\left(R_{t+1}+\g\th_t\tr\ph_{t+1}-\th_t\tr\ph_t\right)\ph_t \\
  &= \th_t + \a\Big(\underbrace{F_t\rho_t R_{t+1}\ph_t}_{\bb_t} - \underbrace{F_t\rho_t\ph_t\left(\ph_t-\g\ph_{t+1}\right)\tr}_{\A_t}\th_t\Big) \nonumber
\end{align}

Emphatic TD(0)'s $\A$ matrix is
\begin{align}
\A = \lim_{t\ra\infty}\E{\A_t} 
 &= \lim_{t\ra\infty}\Emu{F_t\rho_t\ph_t\left(\ph_t-\g\ph_{t+1}\right)\tr} \nonumber\\
 &= \sum_s d_\mu(s)\lim_{t\ra\infty}\CEmu{F_t\rho_t\ph_t\left(\ph_t-\g\ph_{t+1}\right)\tr}{S_t=s} \nonumber\\ 
 &= \sum_s d_\mu(s)\lim_{t\ra\infty}\CEmu{F_t}{S_t=s}\CEmu{\rho_t\ph_t\left(\ph_t-\g\ph_{t+1}\right)\tr}{S_t=s} \nonumber\\ 
 \noalign{\text{(because, given $S_t$, $F_t$ is independent of $\rho_t\ph_t\left(\ph_t-\g\ph_{t+1}\right)\tr$)}}
 &= \sum_s \underbrace{d_\mu(s)\lim_{t\ra\infty}\CEmu{F_t}{S_t=s}}_{f(s)}\CEmu{\rho_k\ph_k\left(\ph_k-\g\ph_{k+1}\right)\tr}{S_k=s} \nonumber\\ 
 &= \sum_s f(s)\CEpi{\ph_k\left(\ph_k-\g\ph_{k+1}\right)\tr}{S_k=s} \tag{by \eqref{eq:Eshift}}\\ 
 &= \sum_s f(s)\,\ph(s)\!\left(\ph(s)-\g\sum_{s'}[\P_\pi]_{ss'}\ph(s')\right)\tr \nonumber\\ 
 &= \PH\tr\F(\I-\g\P_\pi)\PH, \nonumber
\end{align}
where $\F$ is a diagonal matrix with diagonal elements $f(s)\doteq d_\mu(s)\lim_{t\ra\infty}\CEmu{F_t}{S_t\!=\!s}$, which we assume exists. 
As we show later, the vector $\ff\in\Re^N$ with components $[\ff]_s\doteq f(s)$ can be written as
\begin{align}
 \ff &= \dd_\mu + \g\Ppi\tr\dd_\mu + \left(\g\Ppi\tr\right)^2\dd_\mu + \cdots\label{eq:funrolled}\\
     &= \left(\I - \g\Ppi\tr\right)^{-1} \dd_\mu. \label{eq:f}
\end{align}

The key matrix is $\F\left(\I-\g\Ppi\right)$, and the vector of its column sums is
\begin{align*}
 \1\tr\F(\I-\g \Ppi)
  &= \ff\tr(\I-\g \Ppi) \\
  &= \dd_\mu\tr(\I - \g\Ppi)^{-1}(\I-\g \Ppi) \tag{using \eqref{eq:f}}\\
  &= \dd_\mu\tr,  
\end{align*}
all components of which are positive.
Thus, the key matrix and the $\A$ matrix are positive definite and the algorithm is stable. Emphatic TD(0) is the simplest TD algorithm with linear function approximation proven to be stable under off-policy training.

The \thtwoth example presented earlier (Figure~\ref{fig:th2thcontinuing}) provides some insight into how replacing $\D$ by $\F$ changes the key matrix to make it positive definite. In general, $\ff$ is the expected number of time steps that would be spent in each state during an excursion starting from the behavior distribution $\dd_\mu$. From \eqref{eq:funrolled}, it is $\dd_\mu$ plus where you would get to in one step from $\dd_\mu$, plus where you would get to in two steps, etc., with appropriate discounting. In the example, excursions under the target policy take you to the second state ($2\thet$) and leave you there. Thus you are only in the first state ($\thet$) if you start there, and only for one step, so $f(1)=d_\mu(1)=0.5$. For the second state, you can either start there, with probability $0.5$, or you can get there on the second step (certain except for discounting), with probability $0.9$, or on the third step, with probability $0.9^2$, etc, so $f(2)=0.5+0.9+0.9^2+0.9^3+\cdots= 0.5 + 0.9\cdot 10 =9.5$. Thus, the key matrix is now 
\[
\F(\I-\g\Ppi) 
= \begin{bmatrix} 0.5 & 0 \\ 0 & 9.5 \end{bmatrix}
\times \begin{bmatrix} 1 & \sm0.9 \\ 0 & 0.1 \end{bmatrix}
= \begin{bmatrix} 0.5 & \sm0.45 \\ 0 & 0.95 \end{bmatrix}.
\]
Note that because $\F$ is a diagonal matrix, its only effect is to scale the rows. 
Here it emphasizes the lower row by more than a factor of 10 compared to the upper row, thereby causing the key matrix to have positive column sums and be positive definite (cf.~\eqref{eq:keyoff}). The $\F$ matrix emphasizes the second state, which would occur much more often under the target policy than it does under the behavior policy.

\section{The General Case}

We turn now to a very general case of off-policy learning with linear function approximation. The objective is still to evaluate a policy $\pi$ from a single trajectory under a different  policy $\mu$, but now the value of a state is defined not with respect to a constant discount rate $\g\in[0,1]$, but with respect to a discount rate that varies from state to state according to a \emph{discount function} $\g:\SS\ra[0,1]$ such that $\prod_{k=1}^\infty \g(S_{t+k})=0, \text{w.p.1}, \forall t$. That is, our approximation is still defined by \eqref{eq:vpi}, but now \eqref{eq:G} is replaced by
\begin{align}
G_t 
   &\doteq R_{t+1} + \g(S_{t+1}) R_{t+2} + \g(S_{t+1})\g(S_{t+2}) R_{t+3} + \cdots. 
\end{align}
State-dependent discounting specifies a temporal envelope within which received rewards are accumulated. If $\g(S_k)=0$, then the time of accumulation is fully terminated at step $k>t$, and if $\g(S_k)<1$, then it is partially terminated. We call both of these \emph{soft termination} because they are like the termination of an episode, but the actual trajectory is not affected. Soft termination ends the accumulation of rewards into a return, but the state transitions continue oblivious to the termination.
Soft termination with state-dependent termination is essential for learning models of options (Sutton et al.\ 1999) and other applications.

\begin{figure}
\centerline{\includegraphics[height=0.84in]{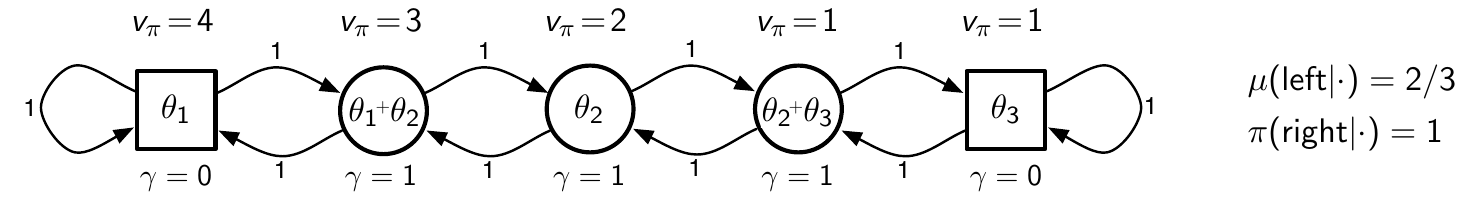}}
\vspace*{-2pt}
\caption{A 5-state chain MDP with soft-termination states at each end.}
\label{fig:5MDP}
\end{figure}

Soft termination is particularly natural in the excursion setting, where it makes it easy to define excursions of finite and definite duration. For example, consider the deterministic MDP shown in Figure~\ref{fig:5MDP}.
There are five states, three of which do not discount at all, $\g(s)=1$, and are shown as circles, and two of which cause complete soft termination, $\g(s)=0$, and are shown as squares. The terminating states do not end anything other than the return; actions are still selected in them and, dependent on the action selected, they transition to next states indefinitely without end. In this MDP there are two actions, $\lef$ and $\righ$, which deterministically cause transitions to the left or right except at the edges, where there may be a self transition. The reward on all transitions is $+1$. The behavior policy is to select \lef $2/3$rds of the time in all states, which causes more time to be spent in states on the left than on the right. The stationary distribution can be shown to be $\dd_\mu \approx (0.52, 0.26, 0.13, 0.06, 0.03)\tr$; more than half of the time steps are spent in the leftmost terminating state.

Consider the target policy $\pi$ that selects the \righ action from all states. The correct value $\vpi(s)$ of each state $s$ is written above it in the figure. For both of the two rightmost states, the right action results in a reward of 1 and an immediate termination, so their values are both 1. For the middle state, following $\pi$ (selecting \righ repeatedly) yields two rewards of 1 prior to termination. There is no discounting ($\g\!=\!1$) prior to termination, so the middle state's value is 2, and similarly the values go up by 1 for each state to its left, as shown. These are the correct values. The approximate values depend on the parameter vector $\th_t$ as suggested by the expressions shown inside each state in the figure. These expressions use the notation $\thet_i$ to denote the $i$th \emph{component} of the current parameter vector $\th_t$. In this example, there are five states and only three parameters, so it is unlikely, and indeed impossible, to represent $\vpi$ exactly. We will return to this example later in the paper.

In addition to enabling definitive termination, as in this example, state-dependent discounting enables a much wider range of predictive questions to be expressed in the form of a value function (Sutton et al.\ 2011, Modayil, White \& Sutton 2014, Sutton, Rafols \& Koop 2006), including option models (Sutton, Precup \& Singh 1999, Sutton 1995). For example, with state-dependent discounting one can formulate questions both about what will happen during a way of behaving and what will be true at its end. A general representation for predictions is a key step toward the goal of representing world knowledge in verifiable predictive terms (Sutton 2009, 2012). The general form is also useful just because it enables us to treat uniformly many of the most important episodic and continuing special cases of interest.

A second generalization, developed for the first time in this paper, is to explicitly specify the states at which we are most interested is obtaining accurate estimates of value. Recall that in parametric function approximation there are typically many more states than parameters ($N\gg n$), and thus it is usually not possible for the value estimates at all states to be exactly correct. Valuing some states more accurately usually means valuing others less accurately, at least asymptotically. In the tabular case where much of the theory of reinforcement learning originated, this tradeoff is not an issue because the estimates of each state are independent of each other, but with function approximation it is necessary to specify relative interest in order to make the problem well defined. Nevertheless, in the function approximation case little attention has been paid in the literature to specifing the relative importance of different states (an exception is Thomas 2014), though there are intimations of this in the initiation set of options (Sutton et al.\ 1999). In the past it was typically assumed that we were interested in valuing states in direct proportion to how often they occur, but this is not always the case. For example, in episodic problems we often care primarily about the value of the first state, or of earlier states generally (Thomas 2014). Here we allow the user to specify the relative interest in each state with a nonnegative \emph{interest function} $i:\SS\ra[0,\infty)$. Formally, our objective is to minimize the Mean Square Value Error (MSVE) with states weighted both by how often they occur and by our interest in them:
\beq
  \MSVE(\th) \doteq \sum_{s\in\SS} d_\mu(s)i(s)\Big(\vpi(s)-\th\tr\ph(s)\Big)^2. \label{eq:MSVE}
\eeq
For example, in the 5-state example in Figure~\ref{fig:5MDP}, we could choose $i(s)=1, \forall s\in\SS$, in which case we would be primarily interested in attaining low error in the states on the left side, which are visited much more often under the behavior policy. If we want to counter this, we might chose $i(s)$ larger for states toward the right. Of course, with parametric function approximation we presumably do not have access to the states as individuals, but certainly we could set $i(s)$ as a function of the features in $s$. In this example, choosing $i(s)=1+\phix_2(s) + 2\phix_3(s)$ (where $\phix_i(s)$ denotes the $i$th component of $\ph(s)$) would shift the focus on accuracy to the states on the right, making it substantially more balanced.

The third and final generalization that we introduce in this section is general bootstrapping. Conventional TD\la uses a bootstrapping parameter $\l\in[0,1]$; we generalize this to a \emph{bootstrapping function} $\l:\SS\ra[0,1]$ specifying a potentially different degree of bootstrapping, $1-\l(s)$, for each state $s$. 
General bootstrapping of this form has been partially developed in several previous works (Sutton 1995, Sutton \& Barto 1998, Maei \& Sutton 2010, Sutton et al.~2014, cf.~Yu 2012).
As a notational shorthand, let us use $\l_t\doteq \l(S_t)$ and $\g_t\doteq \g(S_t)$. 
Then we can define a general notion of bootstrapped return, the $\l$-return with state-dependent bootstrapping and discounting:
\begin{align}
 \Gl_t   
   &\doteq  R_{t+1} + \g_{t+1} \left[(1-\l_{t+1}) \th_t\tr\ph_{t+1}
      + \l_{t+1} \Gl_{t+1}\right].             
\end{align}
The $\l$-return plays a key role in the theoretical understanding of TD methods, in particular, in their forward views (Sutton \& Barto 1998, Sutton, Mahmood, Precup \& van Hasselt 2014). In the forward view, $\Gl_t$ is thought of as the target for the update at time $t$, even though it is not available until many steps later (when complete termination $\g(S_k)=0$ has occurred for the first time for some $k>t$).

Given these generalizations, we can now specify our final new algorithm, \emph{emphatic TD\la}, by the following four equations, for $t\ge 0$:
\begin{align}
 \th_{t+1} &\doteq  \th_t + \a\left(R_{t+1} + \g_{t+1}\th_t\tr\ph_{t+1} - \th_t\tr\ph_t\right)\ee_t  \label{eq:tht}\\
     \ee_t &\doteq  \rho_t\left(\g_t\l_t\ee_{t-1} + M_t\ph_t \right),\text{~~~~with~}\ee_{-1}\doteq \bm 0\label{eq:et}\\
       M_t &\doteq   \l_t \, i(S_t) + (1-\l_t) F_t \label{eq:Mt}\\
       F_t &\doteq  \rho_{t-1}\g_t F_{t-1} + i(S_t), \text{~~~~with~}F_0\doteq i(S_0),  \label{eq:Ft}
\end{align}
where $F_t\ge 0$ is a scalar memory called the \emph{followon trace}. The quantity $M_t\ge 0$ is termed the \emph{emphasis} on step $t$.

\section{Off-policy Stability of Emphatic TD\la}\label{sec:off_policy}
\label{sec:ETDanalysis}

As usual, to analyze the stability of the new algorithm we examine its $\A$ matrix. 
The stochastic update can be written:
\begin{align}  
\th_{t+1} 
 &\doteq  \th_t + \a\left(R_{t+1} + \g_{t+1}\th_t\tr\ph_{t+1} - \th_t\tr\ph_t\right)\ee_t\nonumber\\
  &= \th_t + \a\Big(\underbrace{\ee_t R_{t+1}}_{\bb_t} - \underbrace{\ee_t\left(\ph_t-\g_{t+1}\ph_{t+1}\right)\tr}_{\A_t}\th_t\Big).  \nonumber
\end{align}
Thus, 
\begin{align}
\A = \lim_{t\ra\infty}\E{\A_t}
 &= \lim_{t\ra\infty}\Emu{\ee_t\left(\ph_t-\g_{t+1}\ph_{t+1}\right)\tr} \nonumber\\
 &= \sum_s d_\mu(s)\lim_{t\ra\infty}\CEmu{\ee_t\left(\ph_t-\g_{t+1}\ph_{t+1}\right)\tr}{S_t\!=\!s} \nonumber\\ 
 &= \sum_s d_\mu(s)\lim_{t\ra\infty}\CEmu{\rho_t\left(\g_t\l_t\ee_{t-1} + M_t\ph_t \right)\left(\ph_t-\g_{t+1}\ph_{t+1}\right)\tr}{S_t\!=\!s} \nonumber\\ 
 &= \sum_s d_\mu(s)\lim_{t\ra\infty}\CEmu{\left(\g_t\l_t\ee_{t-1} + M_t\ph_t \right)}{S_t\!=\!s}\CEmu{\rho_t(\ph_t-\g_{t+1}\ph_{t+1})\tr}{S_t\!=\!s} \nonumber\\ 
 \noalign{\text{(because, given $S_t$, $\ee_{t-1}$ and $M_t$ are independent of $\rho_t(\ph_t-\g_{t+1}\ph_{t+1})\tr$)}}
 &= \sum_s \underbrace{d_\mu(s)\lim_{t\ra\infty}\CEmu{\left(\g_t\l_t\ee_{t-1} + M_t\ph_t \right)}{S_t\!=\!s}}_{\ee(s)\in\Re^n}\CEmu{\rho_k(\ph_k-\g_{k+1}\ph_{k+1})\tr}{S_k\!=\!s} \nonumber\\ 
 &= \sum_s \ee(s)\CEpi{\ph_k-\g_{k+1}\ph_{k+1}}{S_k\!=\!s}\tr \tag{by \eqref{eq:Eshift}}\\ 
 &= \sum_s \ee(s)\left(\ph(s)-\sum_{s'}[\P_\pi]_{ss'}\g(s')\ph(s')\right)\tr \nonumber\\ 
 &= \Em(\I-\P_\pi\G)\PH, \label{eq:AfromE}
\end{align}
where $\Em$ is an $N\times n$ matrix $\Em\tr\doteq [\ee(1), \cdots, \ee(N)]$, and 
$\ee(s)\in\Re^n$ is defined by\footnote{Note that this is a slight abuse of notation; $\ee_t$ is a vector random variable, one per time step, and $\ee(s)$ is a vector expectation, one per state.}: 
\begin{align*}
\ee(s) &\doteq d_\mu(s)\lim_{t\ra\infty}\CEmu{\g_t\l_t\ee_{t-1} + M_t\ph_t}{S_t\!=\!s}\tag{assuming this exists}\\
      &=\underbrace{d_\mu(s)\lim_{t\ra\infty}\CEmu{M_t}{S_t\!=\!s}}_{m(s)}\ph(s) + \g(s)\l(s)d_\mu(s)\lim_{t\ra\infty}\CEmu{\ee_{t-1}}{S_t\!=\!s}\\
      &=m(s)\ph(s) \!+\! \g(s)\l(s)d_\mu(s)\!\lim_{t\ra\infty}\sum_{\so,\ao} \CP{S_{t-1}\!\!=\!\so,A_{t-1}\!\!=\!\ao}{S_t\!\!=\!\!s} \CEmu{\ee_{t-1}}{S_{t-1}\!\!=\!\so,A_{t-1}\!\!=\!\ao}\nonumber\\
      &=m(s)\ph(s) + \g(s)\l(s)d_\mu(s)\sum_{\so,\ao} \frac{d_\mu(\so)\mu(\ao|\so)p(s|\so,\ao)}{d_\mu(s)}\lim_{t\ra\infty}\CEmu{\ee_{t-1}}{S_{t-1}\!=\!\so,A_{t-1}\!=\!\ao}\nonumber\\
\noalign{\text{(using the definition of a conditional probability, a.k.a.\ Bayes rule)}}
      &=\!m(s)\ph(s) \!+\! \g(s)\l(s)\!\!\sum_{\so,\ao} \!d_\mu(\so)\mu(\ao|\so)p(s|\so,\ao)\!\frac{\pi(\ao|\so)}{\mu(\ao|\so)}\!\lim_{t\ra\infty}\!\CEmu{\g_{t-1}\l_{t-1}\ee_{t-2}\!+\!M_{t-1}\ph_{t-1}}{S_{t-1}\!\!=\!\!\so}\nonumber\\
      &=m(s)\ph(s) + \g(s)\l(s)\sum_{\so}\left(\sum_{\ao} \pi(\ao|\so)p(s|\so,\ao) \right)\ee(\so)\\
      &=m(s)\ph(s) + \g(s)\l(s)\sum_{\so}[\Ppi]_{\so s} \ee(\so). 
\end{align*}
We now introduce three $N\times N$ diagonal matrices: $\M$, which has the $m(s)\doteq d_\mu(s)\lim_{t\ra\infty}$ $\CEmu{M_t}{S_t\!=\!s}$ on its diagonal; $\G$, which has the $\g(s)$ on its diagonal; and $\L$, which has the $\l(s)$ on its diagonal. With these we can write the equation above entirely in matrix form, as
\begin{align*}
\Em\tr&=\PH\tr\M + \Em\tr\Ppi\G\L\\
      &=\PH\tr\M + \PH\tr\M\Ppi\G\L + \PH\tr\M(\Ppi\G\L)^2 + \cdots\\
      &= \PH\tr\M(\I -\Ppi\G\L)^{-1}. 
\end{align*}
Finally, combining this equation with \eqref{eq:AfromE} we obtain
\beq
 \A = \PH\tr\M(\I -\Ppi\G\L)^{-1}(\I-\Ppi\G)\PH, \label{eq:ETDA}
\eeq
and through similar steps one can also obtain emphatic TD\la's $\bb$ vector, 
\beq
\bb=\Em\rr_\pi=\PH\tr\M(\I -\Ppi\G\L)^{-1}\rr_\pi, \label{eq:b}
\eeq
where $\rr_\pi$ is the $N$-vector of expected immediate rewards from each state under $\pi$.

Emphatic TD\la's key matrix, then, is $\M(\I -\Ppi\G\L)^{-1}(\I-\Ppi\G)$. To prove that it is positive definite we will follow the same strategy as we did for emphatic TD(0). The first step will be to write the last part of the key matrix in the form of the identity matrix minus a probability matrix. 
To see how this can be done, consider a slightly different setting in which actions are taken according to $\pi$, and in which $1-\g(s)$ and $1-\l(s)$ are considered probabilities of ending by terminating or by bootstrapping, respectively. That is, for any starting state, a trajectory involves a state transition according to $\Ppi$, possibly terminating according to $\I-\G$, then possibly ending with a bootstrapping event according to $\I-\L$, and then, if neither of these occur, continuing with another state transition and more chances to end, and so on until an ending of one of the two kinds occurs. For any start state $i\in\SS$, consider the probability that the trajectory ends in state $j\in\SS$ with an ending event of the bootstrapping kind (according to $\I-\L$). Let $\Pl$ be the matrix with this probability as its $ij$th component. This matrix can be written
\begin{align*}
 \Pl &= \Ppi\G(\I-\L) + \Ppi\G\L\Ppi\G(\I-\L) + \Ppi\G(\L\Ppi\G)^2(\I-\L) + \cdots \nonumber\\
    &= \bigg(\sum_{k=0}^\infty (\Ppi\G\L)^k\bigg) \Ppi\G(\I-\L) \nonumber\\
    &= (\I-\Ppi\G\L)^{-1} \Ppi\G(\I-\L).\nonumber\\
    &= (\I-\Ppi\G\L)^{-1} (\Ppi\G-\Ppi\G\L)\nonumber\\
    &= (\I-\Ppi\G\L)^{-1} (\Ppi\G-\I+\I-\Ppi\G\L)\nonumber\\
    &= \I - (\I-\Ppi\G\L)^{-1} (\I-\Ppi\G),  
\end{align*}
or, 
\begin{equation}
\I-\Pl = (\I -\Ppi\G\L)^{-1}(\I-\Ppi\G).  \label{eq:Pl}
\end{equation}
It follows then that
$
 \M(\I -\Pl) = \M(\I -\Ppi\G\L)^{-1}(\I-\Ppi\G)  
$ 
is another way of writing emphatic TD\la's key matrix (cf.~\eqref{eq:ETDA}).
This gets us considerably closer to our goal of proving that the key matrix is positive definite. It is now immediate that its diagonal entries are nonnegative and that its off diagonal entries are nonpositive. It is also immediate that its row sums are nonnegative. 

There remains what is typically the hardest condition to satisfy: that the column sums of the key matrix are positive. To show this we have to analyze $\M$, and to do that we first analyze the $N$-vector $\ff$ with components $f(s)\doteq d_\mu(s)\lim_{t\ra\infty}\CEmu{F_t}{S_t\!=\!s}$ (we assume that this limit and expectation exist).
Analyzing $\ff$ will also pay the debt we incurred in Section 4 when we claimed without proof that $\ff$ (in the special case treated in that section) was as given by \eqref{eq:f}. In the general case:
\begin{flalign*}
f(s) &= d_\mu(s) \lim_{t\ra\infty}\CEmu{F_t}{S_t\!=\!s} \\
     &= d_\mu(s) \lim_{t\ra\infty}\CEmu{i(S_t)+\rho_{t-1}\g_t F_{t-1}}{S_t\!=\!s} \tag{by \eqref{eq:Ft}}\\
     &= d_\mu(s)i(s) + d_\mu(s)\g(s)\! \lim_{t\ra\infty}\sum_{\so,\,\ao} \CP{S_{t-1}\!=\!\so,A_{t-1}\!=\!\ao}{S_t\!=\!s} \frac{\pi(\ao|\so)}{\mu(\ao|\so)}\CEmu{F_{t-1}}{S_{t-1}\!=\!\so} \nonumber\\     
     &= d_\mu(s)i(s) + d_\mu(s)\g(s) \sum_{\so,\,\ao} \frac{d_\mu(\so)\mu(\ao|\so)p(s|\so,\ao)}{d_\mu(s)}\frac{\pi(\ao|\so)}{\mu(\ao|\so)} \lim_{t\ra\infty}\CEmu{F_{t-1}}{S_{t-1}\!=\!\so}\\
\noalign{\text{(using the definition of a conditional probability, a.k.a.\ Bayes rule)}}
     &= d_\mu(s)i(s) + \g(s)\sum_{\so,\,\ao} \pi(\ao|\so)p(s|\so,\ao)d_\mu(\so)\lim_{t\ra\infty}\CEmu{F_{t-1}}{S_{t-1}\!=\!\so}\\
     &= d_\mu(s)i(s) + \g(s)\sum_{\so} [\Ppi]_{\so s}f(\so). 
\end{flalign*}
This equation can be written in matrix-vector form, letting $\ii$ be the $N$-vector with components $[\ii]_s \doteq d_\mu(s)i(s)$:
\begin{align}
 \ff &= \ii + \G\Ppi\tr\ff \nonumber\\
     &= \ii + \G\Ppi\tr\ii + (\G\Ppi\tr)^2\ii + \cdots \nonumber\\
     &= \left(\I-\G\Ppi\tr\right)^{-1}\ii.\label{eq:fvec}
\end{align}
This proves \eqref{eq:f}, because there $i(s)\doteq 1, \forall s$ (thus $\ii=\dd_\mu$), and $\g(s)\doteq\g, \forall s$.

We are now ready to analyze $\M$, the diagonal matrix with the $m(s)$ on its diagonal:
\begin{align*}
 m(s) &= d_\mu(s) \lim_{t\ra\infty}\CEmu{M_t}{S_t\!=\!s} \\
      &= d_\mu(s) \lim_{t\ra\infty}\CEmu{\l_t \, i(S_t) + (1-\l_t) F_t}{S_t\!=\!s} \tag{by \eqref{eq:Mt}}\\
      &= d_\mu(s) \l(s)i(s) + \left(1-\l(s)\right) f(s), 
\end{align*}
or, in matrix-vector form, letting $\mm$ be the $N$-vector with components $m(s)$, 
\begin{align}
 \mm &= \L\ii + (\I-\L) \ff \nonumber\\
     &= \L\ii + (\I-\L) \left(\I-\G\Ppi\tr\right)^{-1}\ii 
     \tag{using \eqref{eq:fvec}}\\
     &= \left[\L(\I-\G\Ppi\tr) + (\I-\L)\right](\I-\G\Ppi\tr)^{-1}\ii\nonumber\\
     &= \left(\I-\L\G\Ppi\tr\right)\left(\I - \G\Ppi\tr\right)^{-1}\ii \label{eq:m}\\
     &= \left(\I - \Pl\tr\right)^{-1}\ii. ~~~~~~~~~~~~~~~~~~~~~\tag{using \eqref{eq:Pl}}
\end{align}

Now we are ready for the final step of the proof, showing that all the columns of the key matrix $\M(\I -\Pl)$ sum to a positive number. Using the result above, the vector of column sums is
\begin{align*}
 \1\tr\M(\I-\Pl) 
  &= \mm\tr(\I-\Pl) \\
  &= \ii\tr(\I - \Pl)^{-1}(\I-\Pl) \\
  &= \ii\tr.
\end{align*}
If we further assume that $i(s)>0, \forall s\in\SS$, then the column sums are all positive, the key matrix is positive definite, and emphatic TD\la and its expected update are stable. 
This result can be summarized in the following theorem, the main result of this paper, which we have just proved:

\begin{theorem}[Stability of Emphatic TD\la]
For any
\begin{itemize}
\item
Markov decision process $\{S_t, A_t, R_{t+1}\}_{t=0}^\infty$ with finite state and actions sets $\SS$ and $\AA$,
\item
behavior policy $\mu$ with a stationary invariant distribution $d_\mu(s)>0, \forall s\in\SS$,  
\item
target policy $\pi$ with coverage, i.e., s.t., if $\pi(a|s)>0$, then $\mu(a|s)>0$, 
\item 
discount function $\g:\SS\ra[0,1]$ s.t.\ $\prod_{k=1}^\infty \g(S_{t+k})=0,\text{w.p.1}, \forall t>0$, 
\item
bootstrapping function $\l:\SS\ra[0,1]$, 
\item
interest function $i:\SS\ra(0,\infty)$, 
\item
feature function $\ph:\SS\ra\Re^n$ s.t.\ the matrix $\PH\in\Re^{|\SS|\times n}$ with the $\ph(s)$ as its rows has linearly independent columns, 
\end{itemize}
the $\A$ matrix of linear emphatic TD\la (as given by (\ref{eq:tht}--\ref{eq:Ft}), and assuming the existence of $\lim_{t\ra\infty}\CE{F_t}{S_t\!=s}$ and $\lim_{t\ra\infty}\CE{\ee_t}{S_t\!=s}$,  $\forall s\in\SS$), 
\beq
 \A = \lim_{t\ra\infty}\Emu{\A_t} = \lim_{t\ra\infty}\Emu{\ee_t\left(\ph_t-\g_{t+1}\ph_{t+1}\right)\tr} = \PH\tr\M(\I -\Pl)\PH, \label{eq:A}
\eeq
is positive definite. Thus the algorithm and its expected update are stable. \end{theorem}

\def\Tl{T^{(\l)}}
\def\PBE{\mbox{\rm PBE}}
As mentioned at the outset, stability is necessary but not always sufficient to guarantee convergence of the parameter vector $\th_t$. Yu (in preparation) has recently built on our stability result to show that in fact emphatic TD\la converges with probability one when the step size $\a$ is reduced appropriately over time. 
Convergence as anticipated is to the unique fixed point $\thstar$ of the deterministic algorithm \eqref{eq:Eth}, in other words, to   
\beq 
\A\thstar=\bb \text{~~~~or~~~~}\thstar = \A^{-1}\bb. \label{eq:thinfty}
\eeq
This solution can be characterized as a minimum (in fact, a zero) of the Projected Bellman Error (PBE, Sutton et al.\ 2009) using the $\l$-dependent Bellman operator $\Tl: \Re^N \ra \Re^N$ (Tsitiklis \& Van Roy 1997) and the weighting of states according to their emphasis. For our general case, we need a version of the $\Tl$ operator extended to state-dependent discounting and bootstrapping. This operator looks ahead to future states to the extent that they are bootstrapped from, that is, according to $\Pl$, taking into account the reward received along the way. The appropriate operator, in vector form, is
\beq
 \Tl \vv \doteq  (\I-\Ppi\G\L)^{-1}\rpi + \Pl \vv. \label{eq:Tl}
\eeq
This operator is a contraction with fixed point $\vv=\vpi$.
Recall that our approximate value function is $\PH\th$, and thus the difference between $\PH\th$ and $\Tl(\PH\th)$ is a Bellman-error vector. The projection of this with respect to the feature matrix and the emphasis weighting is the emphasis-weighted PBE:
\begin{align*}
 \PBE(\th) &\doteq \Pi\left(\Tl(\PH\th) - \PH\th\right) \nonumber\\
 &\doteq  \PH(\PH\tr\M\PH)^{-1}\PH\tr\M\left(\Tl(\PH\th) - \PH\th\right)\tag{see Sutton et al.~2009}\\
 &=\PH(\PH\tr\M\PH)^{-1}\PH\tr\M\left((\I-\Ppi\G\L)^{-1}\rpi + \Pl \PH\th - \PH\th\right)\tag{by \eqref{eq:Tl}}\\
 &=\PH(\PH\tr\M\PH)^{-1}\left(\bb + \PH\tr\M(\Pl - \I)\PH\th\right)\tag{by \eqref{eq:b}}\\
 &=\PH(\PH\tr\M\PH)^{-1}\left(\bb - \A\th\right).\tag{by \eqref{eq:A}}
\end{align*}
From \eqref{eq:thinfty}, it is immediate that this is zero at the fixed point $\thstar$, thus $\PBE(\thstar)=0$.
\goodbreak

Finally, let us reconsider our assumption in this section that the interest function $i(s)$ is strictly greater than zero at all states. If the interest were allowed to be zero at some states, then the key matrix would not necessarily be positive definite, but by continuity it seems that it would still have to be positive semi-definite (meaning $\yy\tr\A\yy$ is positive \emph{or zero} for all vectors $\yy$). Semi-definiteness of the key matrix may well be sufficient for most purposes. In particular, we conjecture that it is sufficient to assure convergence (under appropriate step-size conditions) of the estimated values for all states $s$ with nonzero emphasis, $m(s)$
(cf.~Wang \& Bertsekas 2013). A second advantage of a convergence result based on semi-definiteness is that it would presumably also enable removing the artificial assumption that the columns of the feature matrix $\PH$ are linearly independent. 

\section{Derivation of the Emphasis Algorithm}

Emphatic algorithms are based on the idea that if we are updating a state by a TD method, then we should also update each state that it bootstraps from, in direct proportion. For example, suppose we decide to update the estimate at time $t$ with unit emphasis, perhaps because $i(S_t)=1$, and then at time $t+1$ we have $\g(S_{t+1})=1$ and $\l(S_{t+1})=0$. Because of the latter, we are fully bootstrapping from the value estimate at $t+1$ and thus we should also make an update of it with emphasis equal to $t$'s emphasis. 
If instead $\l(S_{t+1})=0.5$, then the update of the estimate at $t+1$ would gain a half unit of emphasis, and the remaining half would still be available to allocate to the updates of the estimate at $t+2$ or later times depending on their $\l$s. And of course there may be some emphasis allocated directly updating the estimate at $t+1$ if $i(S_{t+1})>0$.
Discounting and importance sampling also have effects. At each step $t$, if $\g(S_t)<1$, then there is some degree of termination and to that extent there is no longer any chance of bootstrapping from later time steps. Another way bootstrapping may be cut off is if $\rho_t=0$ (a complete deviation from the target policy). More generally, if $\rho\not= 1$, then the opportunity for bootstrapping is scaled up or down proportionally. 

It may seem difficult to work out precisely how each time step's estimates bootstrap from which later states' estimates for all cases. Fortunately, it has already been done. Equation (6) of the paper by Sutton, Mahmood, Precup, and van Hasselt (2014) specifies this in their ``forward view" of off-policy TD\la with general state-dependent discounting and bootstrapping. From this equation (and their (5)) it is easy to determine the degree to which the update of the value estimate at time $k$ bootstraps from (multiplicatively depends on) the value estimates of each subsequent time $t$. It is
\begin{align*}
\rho_k \left(\prod_{i=k+1}^{t-1} \g_i \l_i \rho_i\right) \g_t (1-\l_t).
\end{align*}
It follows then that the total emphasis on time $t$, $M_t$, should be the sum of this quantity for all times $k<t$, each times the emphasis $M_k$ for those earlier times, plus any intrinsic interest $i(S_t)$ in time $t$:
\begin{align*}
M_t &\doteq i(S_t) + \sum_{k=0}^{t-1} M_k\rho_k \left(\prod_{i=k+1}^{t-1} \g_i \l_i \rho_i\right) \g_t (1-\l_t)  \\
&= \l_ti(S_t) + (1-\l_t)i(S_t) + (1-\l_t) \g_t\sum_{k=0}^{t-1} \rho_k M_k \prod_{i=k+1}^{t-1} \g_i \l_i \rho_i   \\
&= \l_t i(S_t) + (1-\l_t) F_t, 
\end{align*}
which is \eqref{eq:Mt}, where 
\begin{flalign*}
F_{t} 
&\doteq i(S_t) + \g_t\sum_{k=0}^{t-1} \rho_k M_k \prod_{i=k+1}^{t-1} \g_i \l_i \rho_i \\
&= i(S_t) + \g_{t} \left(\rho_{t-1} M_{t-1} +  \sum_{k=0}^{t-2} \rho_k M_k \prod_{i=k+1}^{t-1} \g_i \l_i \rho_i \right) \\
&= i(S_t) + \g_{t} \left(\rho_{t-1} M_{t-1} +  \rho_{t-1}\l_{t-1}\g_{t-1}\sum_{k=0}^{t-2} \rho_k M_k \prod_{i=k+1}^{t-2} \g_i \l_i \rho_i \right) \\
&= i(S_t) + \g_{t} \rho_{t-1}\Bigg(\underbrace{\l_{t-1} i(S_{t-1}) + (1-\l_{t-1})F_{t-1}}_{M_{t-1}} +  \l_{t-1}\g_{t-1}\sum_{k=0}^{t-2} \rho_k M_k \prod_{i=k+1}^{t-2} \g_i \l_i \rho_i    \Bigg) \\
&= i(S_t) + \g_{t} \rho_{t-1}\Bigg(F_{t-1} + \l_{t-1} \bigg({-F_{t-1}} +  \underbrace{i(S_{t-1}) +\g_{t-1}\sum_{k=0}^{t-2} \rho_k M_k \prod_{i=k+1}^{t-2} \g_i \l_i \rho_i}_{F_{t-1}} \bigg)\Bigg) \\
&= i(S_t) + \g_{t} \rho_{t-1}F_{t-1},
\end{flalign*}
which is \eqref{eq:Ft}, completing the derivation of the emphasis algorithm.

\section{Empirical Examples}

In this section we present empirical results with example problems that verify and elucidate the formal results already presented. A thorough empirical comparison of emphatic TD\la with other methods is beyond the scope of the present article.

The main focus in this paper, as in much previous theory of TD algorithms with function approximation, has been on the stability of the expected update. If an algorithm is unstable, as Q-learning and off-policy TD\la are on Baird's (1995) counterexample, then there is no chance of its behaving in a satisfactory manner. 
On the other hand, even if the update is stable it may be of very high variance.
%
Off-policy algorithms involve products of potentially an infinite number of importance-sampling ratios, which can lead to fluxuations of infinite variance.

As an example of what can happen, let's look again at the \thtwoth problem shown in Figure~1 (and shown again in the upper left of Figure~3).
Consider what happens to $F_t$ in this problem if we have interest only in the first state,  and the \righ action happens to be taken on every step (i.e., $i(S_0)=1$ then $i(S_t)=0, \forall t>0$, and $A_{t}=\righ, \forall t\ge0$). In this case, from \eqref{eq:Ft}, 
\begin{align*}
 F_t = \rho_{t-1}\g_t F_{t-1} + i(S_t) 
     = \prod_{j=0}^{t-1} \rho_j \g 
     = (2\cdot 0.9)^t, 
\end{align*}
which of course goes to infinity as $t\ra\infty$. On the other hand, the probability of this specific infinite action sequence is zero, and in fact $F_t$ will rarely take on very high values. In particular, the expected value of $F_t$ remains finite at 
\begin{align*}
\Emu{F_t} &= 0.5 \cdot 2 \cdot 0.9 \cdot \Emu{F_{t-1}} + 0.5 \cdot 0 \cdot 0.9 \cdot \Emu{F_{t-1}}\\
&= 0.9 \cdot \Emu{F_{t-1}}\\
&= 0.9^t, 
\end{align*}
which tends to zero as $t\ra\infty$.
Nevertheless, this problem is indeed a difficult case, as the \emph{variance} of $F_t$ \emph{is} infinite:
\begin{align*}
 \Var{F_t} 
  &= \E{F_t^2} - \left(\E{F_t}\right)^2\\
  &= 0.5^t (2^t 0.9^t)^2 - (0.9^t)^2 \\
  &= (0.9^2 \cdot 2)^t - (0.9^2)^t \\
  &= 1.62^t - 0.81^t, 
\end{align*}
which tends to $\infty$ as $t\ra\infty$. 

\begin{figure}[b]
\centerline{\includegraphics[width=\textwidth]{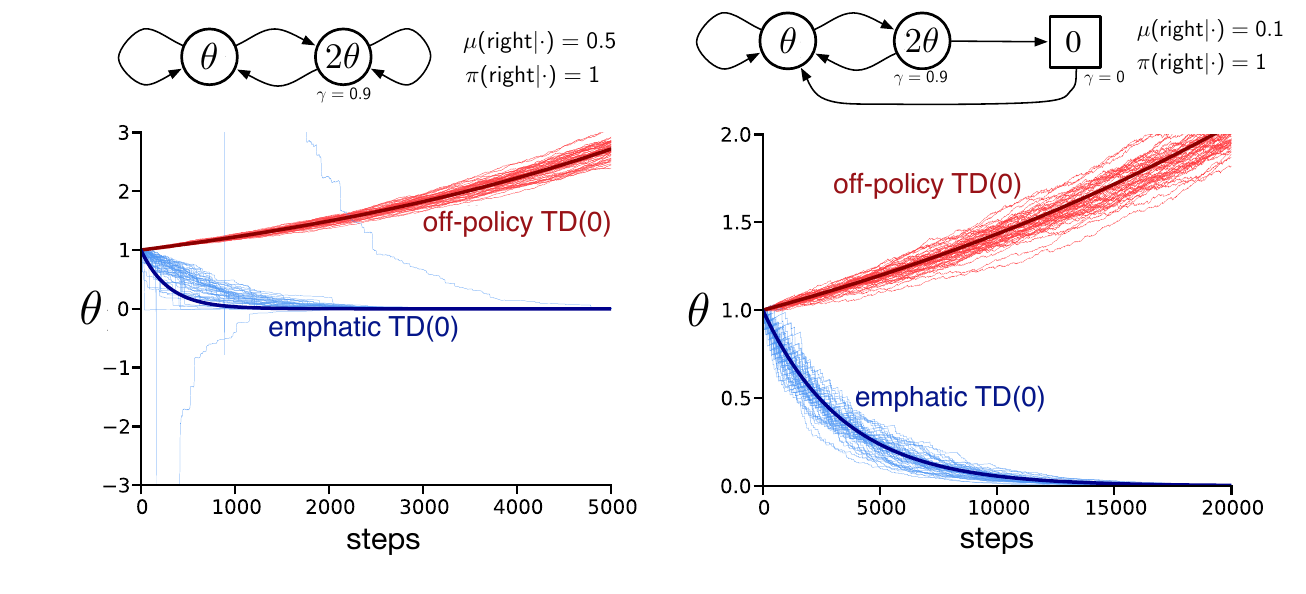}}
\vspace*{-.15in}
\caption{Emphatic TD approaches the correct value of zero, whereas conventional off-policy TD diverges, on fifty trajectories on the \thtwoth problems shown above each graph. Also shown as a thick line is the trajectory of the deterministic expected-update algorithm \eqref{eq:Eth}. On the continuing problem (left) emphatic TD had occasional high variance deviations from zero.}
\end{figure}

So what actually happens on this problem? The thin blue lines in Figure~3~(left) show the trajectories of the single parameter $\thet$ over time in 50 runs with this problem with $\l\!=\!0$ and $\a\!=\!0.001$, starting at $\thet\!=\!1.0$. We see that most trajectories of emphatic TD(0) rapidly approached the correct value of $\thet\!=\!0$, but a few made very large steps away from zero and then returned. Because the variance of $F_t$ (and thus of $M_t$ and $\ee_t$) grows to infinity as $t$ tends to infinity, there is always a small chance of an extremely large fluxuation taking $\thet$ far away from zero. 
Off-policy TD(0), on the other hand, diverged to infinity in all individual runs.

For comparison, Figure~3~(right) shows trajectories for a \thtwoth problem in which $F_t$ and all the other variables and their variances are bounded. In this problem, the target policy of selecting \righ on all steps leads to a soft terminal state ($\g(s)=0$) with fixed value zero, which then transitions back to start again in the leftmost state, as shown in the upper right of the figure. (This is an example of how one can reproduce the conventional notions of terminal state and episode in a soft termination setting.) Here we have chosen the behavior policy to take the action \lef with probability 0.9, so that its stationary distribution distinctly favors the left state, whereas the target policy would spend equal time in each of the two states. This change increases the variance of the updates, so we used a smaller step size, $\alpha=0.0001$; other settings were unchanged. Conventional off-policy TD(0) still diverged in this case, but emphatic TD(0) converged reliably to zero.

\begin{figure}[b]
\centerline{~~~~~~~~~~~~~~\includegraphics[width=4.8in]{5excursion.pdf}}
\centerline{\includegraphics[width=4in]{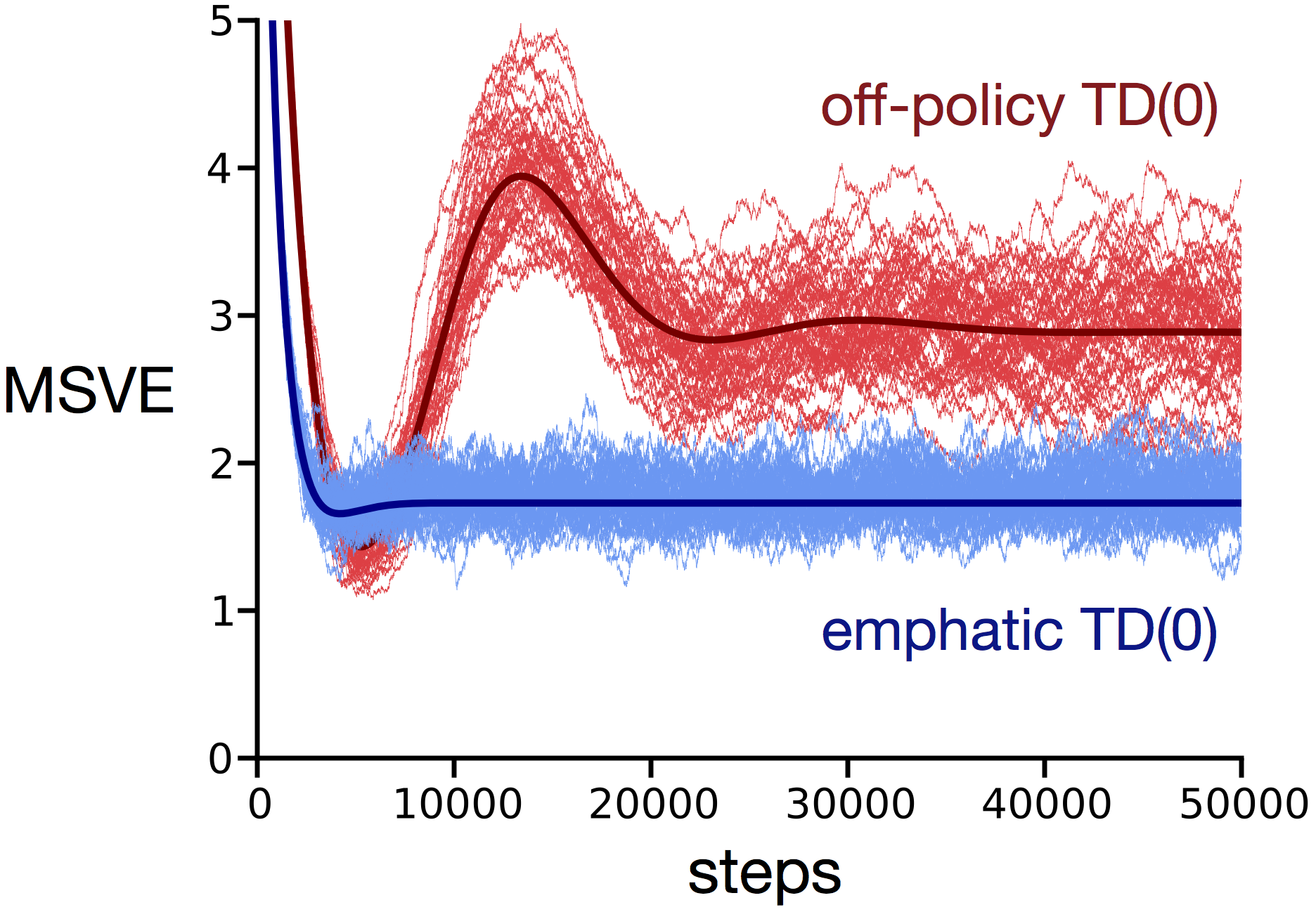}~~~~~~~~~}
\vspace*{-.2in}
 \caption{Twenty learning curves and their analytic expectation on the 5-state problem from Section~5, in which excursions terminate promptly and both algorithms converge reliably. Here $\l=0$, $\th_0=\bm 0$, $\a=0.001$, and $i(s)=1, \forall s$.
The MSVE performance measure is defined in \eqref{eq:MSVE}.}
\end{figure}

Finally, Figure~4 shows trajectories for the 5-state example shown earlier (and again in the upper part of the figure). In this case, everything is bounded under the target policy, and both algorithms converged. The emphatic algorithm achieved a lower MSVE in this example (nevertheless, we do not mean to claim any general empirical advantage for emphatic TD\la at this time).

Also shown in these figures as a thick dark line is the trajectory of the deterministic  algorithm:
$
 \bar\th_{t+1} = \bar\th_t + \a(\bb-\A\bar\th_t)
$ \eqref{eq:Eth}. 
Tsitsiklis and Van Roy (1997) argued that, for small step-size parameters and in the steady-state distribution, on-policy TD\la follows its expected-update algorithm in an ``average" sense, and we see much the same here for emphatic TD\la.

These examples show that although emphatic TD\la is stable for any MDP and all functions $\l$, $\g$ and (positive) $i$, for some problems and functions the parameter vector continues to fluxuate with a chance of arbitrarily large deviations (for constant $\a>0$). It is not clear how great of a problem this is. Certainly it is much less of a problem than the positive instability (Baird 1995) that can occur with off-policy TD\la (stability of the expected update precludes this). The possibility of large fluxuations  may be inherent in any algorithm for off-policy learning using importance sampling with long eligibility traces. 
For example, the updates of GTD\la and GQ\la (Maei 2011) with $\l=1$ will tend to infinite variance as $t\ra\infty$ on Baird's counterexample and on the example in Figures 1 and 3(left).
And, as mentioned earlier, convergence with probability one can still be guaranteed if $\a$ is reduced appropriately over time (Yu, in preparation).


In practice, however, even when asymptotic convergence can be guaranteed, high variance can be problematic as it may require very small step sizes and slow learning.
High variance frequently arises in off-policy algorithms when they are Monte Carlo algorithms (no TD learning) or they have eligibility traces with high $\l$ (at $\l\!=\!1$, TD algorithms become Monte Carlo algorithms). In both cases the root problem is the same: importance sampling ratios that become very large when multiplied together. For example, in the \thtwoth problem discussed at the beginning of this section, the ratio was only two, but the products of successive twos rapidly produced a very large $F_t$. Thus, the first way in which variance can be controlled is to ensure that large products cannot occur. We are actually concerned with products of both $\rho_t$s and $\g_t$s. Occasional termination ($\g_t\!=\!0$), as in the 5-state problem, is thus one reliable way of preventing high variance. Another is through choice of the target and behavior policies that together determine the importance sampling ratios. For example, one could define the target policy to be equal to the behavior policy whenever the followon or eligibility traces exceed some threshold. These tricks can also be done prospectively. White (in preparation) proposed that the learner compute at each step the variance of what GTD\la's traces would be on the following step. If the variance is in danger of becoming too large, then $\l_t$ is reduced for that step to prevent it. For emphatic TD\la, the same conditions could be used to adjust $\g_t$ or one of the policies to prevent the variance from growing too large. Another idea for reducing variance is to use \emph{weighted} importance sampling (as suggested by Precup et al.~2001) together with the ideas of Mahmood et al.~(2014, in preparation) for extending weighted importance sampling to linear function approximation. Finally, a good solution may even be found by something as simple as bounding the values of $F_t$ or $\ee_t$. This would limit variance at the cost of bias, which might be a good tradeoff if done properly.

\section{Conclusions and Future Work}

We have introduced a way of varying the emphasis or strength of the updates of TD learning algorithms from step to step, based on importance sampling, that should result in much lower variance than previous methods (Precup et al.\ 2001). 
In particular, we have introduced the emphatic TD\la algorithm and shown that it  solves the problem of instability that plagues conventional TD\la when applied in off-policy training situations in conjunction with linear function approximation. Compared to gradient-TD methods, emphatic TD\la is simpler in that it has a single parameter vector and a single step size rather than two of each.
The per-time-step complexities of gradient-TD and emphatic-TD methods are both linear in the number of parameters; both are much simpler than quadratic complexity methods such LSTD\la and its off-policy variants.
We have also presented a few empirical examples of emphatic TD(0) compared to conventional TD(0) adapted to off-policy training. These examples illustrate some of emphatic TD\la's basic strengths and weaknesses, but a proper empirical comparison with other methods remains for future work. Extensions of the emphasis idea to action-value and control methods such as Sarsa\la and Q\la, to true-online forms (van Seijen \& Sutton 2014), and to weighted importance sampling (Mahmood et al.~2014, in preparation) are also natural and remain for future work. Yu (in preparation) has recently extended the emphatic idea to a least-squares algorithm and proved it and our emphatic TD\la convergent with probability one.

Two additional ideas for future work deserve special mention. 

First, note that the present work has focused on ways of ensuring that the key matrix is positive definite, which implies positive definiteness of the $\A$ matrix and thus that the update is stable. An alternative strategy would be to work directly with the $\A$ matrix. Recall that the $\A$ matrix is vastly smaller than the key matrix; it has a row and column for each \emph{feature}, whereas the key matrix has a row and column for each \emph{state}. It might be feasible then to  keep statistics for each row and column of $\A$, whereas of course it would not be for the large key matrix. For example, one might try to use such statistics to directly test for diagonal dominance (and thus positive definiteness) of $\A$. If it were possible to adjust some of the free parameters (e.g., the $\l$ or $i$ functions) to ensure positive definiteness while reducing the variance of $F_t$, then a substantially improved algorithm might be found.

The second idea for future work is that the emphasis algorithm, by tracing the dependencies among the estimates at various states, is doing something clever that ought to show up as improved bounds on the asymptotic approximation error. The bound given by Tsitsiklis and Van Roy (1997) probably cannot be significantly improved if $\l$, $\g$, $i$, and $\rho$ are all constant, because in this case emphasis asymptotes to a constant that can be absorbed into the step size. But if any of these vary from step to step, then emphatic TD\la is genuinely different and may improve over conventional TD\la. 
In particular, consider an episodic on-policy case where $i(s)\!\doteq\! 1$ and $\l(s)\!\doteq\! 0$, for all $s\in\SS$, and $\g(s)\!\doteq\! 1$ for all states except for a terminal state where it is zero (and from which a new episode starts).
In this case emphasis would increase linearly within an episode to a maximum on the final state, whereas conventional TD\la would give equal weight to all steps within the episode. If the feature representation were insufficient to represent the value function exactly, then the emphatic algorithm might improve over the conventional algorithm in terms of asyptotic MSVE \eqref{eq:MSVE}. Similarly, improvements in asymptotic MSVE over conventional algorithms might be possible whenever $i$ varies from state to state, such as in the common episodic case in which we are interested only in accurately valuing the start state of the episode, and yet we choose $\l<1$ to reduce variance. There may be a wide range of interesting theoretical and empirical work to be done along these lines.

%

\section*{Acknowledgements}
The authors thank Hado van Hasselt, Doina Precup, Huizhen Yu, and Brendan Bennett for insights and discussions contributing to the results presented in this paper, and the entire Reinforcement Learning and Artificial Intelligence research group for providing the environment to nurture and support this research. We gratefully acknowledge funding from Alberta Innovates -- Technology Futures and from the Natural Sciences and Engineering Research Council of Canada.

\section*{References}

\parindent=0pt
\def\hangin{\hangindent=0.15in}
\parskip=6pt

\hangin
Baird, L.~C. (1995).
\newblock Residual algorithms: {R}einforcement learning with function
  approximation.
\newblock In {\em Proceedings of the
  12th International Conference on Machine Learning}, pp.~30--37.
  Morgan Kaufmann, San Francisco. Important modifications and errata added to the online version on November 22, 1995.


\hangin
Bertsekas, D.~P. (2012).
\newblock {\em Dynamic Programming and Optimal Control: Approximate Dynamic Programming}, Fourth Edition.
\newblock Athena Scientific, Belmont, MA.

\hangin
Bertsekas, D.~P., Tsitsiklis, J.~N. (1996).
\newblock {\em Neuro-Dynamic Programming}.
\newblock Athena Scientific, Belmont, MA.

\hangin
Boyan, J.~A., (1999).
Least-squares temporal difference learning.
In \emph{Proceedings of the 16th International Conference on Machine Learning}, pp. 49--56. 

\hangin
Bradtke, S., Barto, A. G. (1996).
\newblock Linear least-squares algorithms for temporal difference learning.
\newblock {\em Machine Learning 22}:33--57.

\hangin
Dayan, P. (1992).
\newblock The convergence of {TD}($\lambda$) for general $\lambda$.
\newblock {\em Machine Learning 8}:341--362.


\hangin
Dann, C., Neumann, G., Peters, J. (2014).
Policy evaluation with temporal differences: A survey and comparison.
\emph{Journal of Machine Learning Research 15}:809--883.

\hangin
Geist, M., Scherrer, B. (2014).
Off-policy learning with eligibility traces: A survey.
\emph{Journal of Machine Learning Research 15}:289--333.

\hangin
Gordon, G.~J. (1995).
\newblock Stable function approximation in dynamic programming.
\newblock In A.~Prieditis and S.~Russell (eds.), {\em Proceedings of the
  12th International Conference on Machine Learning}, pp.~261--268.
   Morgan Kaufmann, San Francisco.
\newblock An expanded version was published as Technical Report CMU-CS-95-103.
  Carnegie Mellon University, Pittsburgh, PA, 1995.

\hangin
Gordon, G.~J. (1996).
\newblock Stable fitted reinforcement learning.
\newblock In D.~S.~Touretzky, M.~C.~Mozer, M.~E.~Hasselmo (eds.), {\em
Advances in
  Neural Information Processing Systems: {P}roceedings of the 1995 Conference},
  pp.~1052--1058. MIT Press, Cambridge, MA.

\hangin
Hackman, L. (2012).
\emph{Faster Gradient-TD Algorithms}.
MSc thesis, University of Alberta.

\hangin
Klopf, A. H. (1988). A neuronal model of classical conditioning. \emph{Psychobiology 16}(2):85--125.

\hangin
Kolter, J. Z. (2011). The fixed points of off-policy TD. In \emph{Advances in Neural Information Processing Systems 24}, pp.~2169--2177.

\hangin
Lagoudakis, M., Parr, R. (2003). Least squares policy iteration. 
\textit{Journal of Machine Learning Research 4}:1107--1149.

\hangin
Ludvig, E. A., Sutton, R. S., Kehoe, E. J. (2012). Evaluating the TD model of classical conditioning. \emph{Learning \& behavior 40}(3):305--319.

\hangin
Maei, H.~R. (2011).
\emph{Gradient Temporal-Difference Learning Algorithms}. 
PhD thesis, University of Alberta. 

\hangin
Maei, H.~R., Sutton, R.~S. (2010). 
GQ($\l$): A general gradient algorithm for temporal-difference prediction learning with eligibility traces. 
In  \emph{Proceedings of the Third Conference on Artificial General Intelligence}, pp.~91--96. Atlantis Press.

\hangin
Maei, H. R., Szepesv\'ari, Cs., Bhatnagar, S., Sutton, R. S. (2010). Toward off-policy learning control with function approximation. In \emph{Proceedings of the 27th International Conference on Machine Learning}, pp.~719--726.

\hangin
Mahmood, A. R., van Hasselt, H., Sutton, R. S. (2014).
Weighted importance sampling for off-policy learning with linear function approximation.
\textit{Advances in Neural Information Processing Systems 27}.

\hangin
Mahmood, A. R., Sutton, R. S. (in preparation).
Off-policy learning based on weighted importance sampling with linear computational complexity.

\hangin
Modayil, J., White, A., Sutton, R.~S. (2014). Multi-timescale nexting in a reinforcement learning robot. \emph{Adaptive Behavior 22}(2):146--160.

\hangin
Nedi\'c, A., Bertsekas, D. P. (2003). Least squares policy evaluation algorithms with linear function approximation. \emph{Discrete Event Dynamic Systems 13}(1-2):79--110.

\hangin
Niv, Y., Schoenbaum, G. (2008). Dialogues on prediction errors. \emph{Trends in cognitive sciences 12}(7):265--272.

\hangin
O'Doherty, J. P. (2012). Beyond simple reinforcement learning: The computational neurobiology of reward learning and valuation. \emph{European Journal of Neuroscience 35}(7):987--990.


\hangin
Precup, D., Sutton, R.~S., Dasgupta, S. (2001).
Off-policy temporal-difference learning with function approximation. 
In \emph{Proceedings of the 18th International Conference on Machine Learning}, 
pp.~417--424.

\hangin
Precup, D., Sutton, R.~S., Singh, S. (2000).
Eligibility traces for off-policy policy evaluation. 
In \emph{Proceedings of the 17th International Conference on Machine Learning}, 
pp.~759--766. Morgan Kaufmann.


\hangin
Rummery, G.~A. (1995). 
\emph{Problem Solving with Reinforcement Learning}. 
PhD thesis, University of Cambridge.

\hangin
Samuel, A.~L. (1959).
Some studies in machine learning using the game of checkers.
{\em IBM Journal on Research and Development 3}:210--229.
Reprinted in E.~A.~Feigenbaum, \& J.~Feldman (Eds.),
{\em Computers and thought}.
New York: McGraw-Hill.

\hangin
Schultz, W., Dayan, P., Montague, P. R. (1997). A neural substrate of prediction and reward. \emph{Science 275}(5306):1593--1599.


\hangin
Sutton, R. S. (1988). Learning to predict by the methods of temporal differences. \emph{Machine Learning 3}:9--44, erratum p.~377.

\hangin
Sutton, R.~S. (1995). TD models: Modeling the world at a mixture of time scales. In \emph{Proceedings of the 12th International Conference on Machine Learning}, pp.~531--539. Morgan Kaufmann.

\hangin
Sutton, R.\ S. (2009). 
The grand challenge of predictive empirical abstract knowledge. 
\textit{Working Notes of the IJCAI-09 Workshop on Grand Challenges for Reasoning from Experiences}.

\hangin
Sutton, R.\ S. (2012). 
Beyond reward: The problem of knowledge and data. In \emph{Proceedings of the 21st International Conference on Inductive Logic Programming}, S.~H.\ Muggleton, A. Tamaddoni-Nezhad, F.~A.\ Lisi (Eds.): ILP 2011, LNAI 7207, pp.~2--6. Springer, Heidelberg.


\hangin
Sutton, R.~S., Barto, A.~G. (1990).
\newblock Time-derivative models of Pavlovian reinforcement.
\newblock In M.~Gabriel and J.~Moore (Eds.), {\em Learning and
  Computational Neuroscience: {F}oundations of Adaptive Networks},
  pp.~497--537. MIT Press, Cambridge, MA.

\hangin
Sutton, R. S., Barto, A. G. (1998). \emph{Reinforcement Learning: An Introduction}. MIT Press.

\hangin
Sutton, R.~S., Mahmood, A.~R.,  Precup, D., van Hasselt, H. (2014).
A new Q($\lambda$) with interim forward view and Monte Carlo equivalence.
In \emph{Proceedings of the 31st International Conference on Machine Learning}.
JMLR W\&CP 32(2).

\hangin
Sutton, R.~S., Maei, H.~R., Precup, D.,  Bhatnagar, S., Silver, D., Szepesv{\'a}ri, {Cs}.,  Wiewiora, E. (2009).
Fast gradient-descent methods for temporal-difference learning with linear function approximation.
In \emph{Proceedings of the 26th International Conference on Machine Learning}, pp. 993--1000, ACM.

\hangin
Sutton, R.~S., Modayil, J., Delp, M., Degris, T., Pilarski, P.~M., White, A., Precup, D. (2011). Horde: A scalable real-time architecture for learning knowledge from unsupervised sensorimotor interaction. In \emph{Proceedings of the 10th International Conference on Autonomous Agents and Multiagent Systems}, pp. 761--768.

\hangin
Sutton, R.~S., Precup D., Singh, S. (1999).
\newblock  Between {MDP}s and semi-{MDP}s: A framework for temporal abstraction in reinforcement learning.
\newblock  {\em Artificial Intelligence 112}:181--211.

\hangin
Sutton, R.~S., Rafols, E.~J., Koop, A. (2006). 
Temporal abstraction in temporal-difference networks.  
 \textit{Advances in Neural Information Processing Systems 18}. MIT Press.

\hangin
Tesauro, G. (1992). Practical issues in temporal difference learning. 
\emph{Machine Learning 8}:257--277.

\hangin
Tesauro, G. (1995). Temporal difference learning and TD-Gammon. 
\emph{Communications of t he ACM 38}(3):58--68.

\hangin
Thomas, P. (2014).
Bias in natural actor--critic algorithms.
In \emph{Proceedings of the 31st International Conference on Machine Learning}.
JMLR W\&CP 32(1):441--448.

\hangin
Tsitsiklis, J.~N., {Van Roy}, B. (1996).
\newblock Feature-based methods for large scale dynamic programming.
\newblock {\em Machine Learning}, 22:59--94.

\hangin
Tsitsiklis, J. N., Van Roy, B. (1997). 
An analysis of temporal-difference learning with function approximation. 
{\em IEEE Transactions on Automatic Control 42}:674--690.


\hangin
van Seijen, H., Sutton, R.~S. (2014).
True online TD($\l$).
In \emph{Proceedings of the 31st International Conference on Machine Learning}.
JMLR W\&CP 32(1):692--700.

\hangin
Varga, R.~S. (1962).
{\sl Matrix Iterative Analysis.}
Englewood Cliffs, NJ: Prentice-Hall.

\hangin
Wang, M., Bertsekas, D. P. (2013). Stabilization of stochastic iterative methods for singular and nearly singular linear systems. \emph{Mathematics of Operations Research 39}(1):1-30.

\hangin
Watkins, C. J. C. H. (1989). 
\emph{Learning from Delayed Rewards}. PhD thesis, University of Cambridge.

\hangin
Watkins, C.~J.~C.~H., Dayan, P. (1992).
\newblock Q-learning.
\newblock {\em Machine Learning 8}:279--292.

\hangin
White, A. (in preparation).
\emph{Developing a Predictive Approach to Knowledge}.
Phd thesis, University of Alberta.

\hangin
Yu, H. (2010). Convergence of least squares temporal difference methods under general conditions. In \emph{Proceedings of the 27th International Conference on Machine Learning}, pp.~1207--1214.

\hangin
Yu, H. (2012). Least squares temporal difference methods: An analysis under general conditions. \emph{SIAM Journal on Control and Optimization 50}(6), 3310--3343.

\hangin
Yu, H. (in preparation).
On convergence of emphatic temporal-difference learning.


\end{document}